\pdfoutput=1
\documentclass[11pt]{article}

\usepackage[final]{acl}

\usepackage{times}
\usepackage{latexsym}

\usepackage[T1]{fontenc}
\usepackage[utf8]{inputenc}

\usepackage{microtype}

\usepackage{inconsolata}

\usepackage{graphicx}

\usepackage{booktabs}
\usepackage{multirow}
\usepackage{makecell}
\usepackage{soul}

\newcommand{\rescell}[2]{\makecell{$#1{\scriptstyle \pm#2}$}}
\newcommand{\narrowbotc}[1]{{\colorbox{yellow}{\parbox[t][][t]{23em}{#1}}}}
\newcommand{\change}[1]{#1}

\title{Aggregation Artifacts in Subjective Tasks\\Collapse Large Language Models' Posteriors}

\author{
 \textbf{Georgios Chochlakis\textsuperscript{1}},
 \textbf{Alexandros Potamianos\textsuperscript{1,2}},\\
 \textbf{Kristina Lerman\textsuperscript{1}},
 \textbf{Shrikanth Narayanan\textsuperscript{1}}
\\
 \textsuperscript{1}University of Southern California,
 \textsuperscript{2}National Technical University of Athens
\\
 \small{
   \textbf{Correspondence:} \href{mailto:chochlak@usc.edu}{chochlak@usc.edu}
 }
}

\begin{document}
\maketitle
\begin{abstract}
In-context Learning (ICL) has become the primary method for performing natural language tasks with Large Language Models (LLMs). The knowledge acquired during pre-training is crucial for this \textit{few-shot} capability, providing the model with \textit{task priors}. However, recent studies have shown that ICL predominantly relies on retrieving task priors rather than ``learning'' to perform tasks. This limitation is particularly evident in complex subjective domains such as emotion and morality, where priors significantly influence posterior predictions.
In this work, we examine whether this is the result of the aggregation used in corresponding datasets, where trying to combine low-agreement, disparate annotations might lead to annotation artifacts that create detrimental noise in the prompt.
Moreover, we evaluate the posterior bias towards certain annotators by grounding our study in appropriate, quantitative measures of LLM priors. Our results indicate that aggregation is a \textit{confounding factor} in the modeling of subjective tasks, and advocate focusing on modeling individuals instead. However, aggregation does not explain the entire gap between ICL and the state of the art, meaning other factors in such tasks also account for the observed phenomena.
Finally, by rigorously studying annotator-level labels, we find that it is possible for minority annotators to both better align with LLMs and have their perspectives further amplified.\footnote{code and data available at: \url{https://github.com/gchochla/aggregation-artifacts-llms}}
\end{abstract}

\section{Introduction}

Large Language Models (LLMs) \cite{radford2019language, ouyangTrainingLanguageModels2022, touvronLlamaOpenFoundation2023, dubey2024llama, brownLanguageModelsAre2020, achiam2023gpt} have come to dominate language processing as generalists that can perform many tasks. This dominance comes from the emergence of methods such as In-Context Learning (ICL; \citealt{brownLanguageModelsAre2020}) and Chain-of-Throught prompting (CoT; \citealt{weiChainThoughtPrompting2022}), wherein LLMs perform tasks by leveraging input-output demonstrations and task instructions in the prompt only, without any parameter updates.

While ICL is often contrasted with traditional in-weights learning (i.e., gradient-based updates of the models' parameters) \cite{kossen2023context, chanDataDistributionalProperties2022}, the ICL abilities of LLMs depend on their general, in-weights prior knowledge, allowing them to perform many tasks in a zero-shot or few-shot manner. Therefore, studying how LLMs leverage the context in relation with their existing knowledge is a prerequisite to understanding ICL.

Prior work found evidence that LLMs may be overly reliant on their prior knowledge, disregarding the demonstrations in the prompt. Specifically, \citet{min2022rethinking} demonstrated that, under certain circumstances, LLMs ignore the provided signal in their prompt in the form of the mapping between inputs and outputs, and instead act as a database of tasks; they focus on the examples and the labels \textit{independently} to fetch the underlying task \cite{xie2021explanation}: \citet{min2022rethinking} sampled examples and labels independently, and showed very little change in performance. Since no annotations are provided, the setting is given the status of ``zero-shot'' inference, namely \textbf{task-recognition zero-shot}. While follow-up work~\cite{kossen2023context, wei2023larger, pan2023context, yoo2022ground} has further studied this phenomenon and questioned its generality, more recent work~\cite{chochlakis2024strong, chochlakis2024larger} has provided further, quantitative evidence for the relative importance of prior and evidence in the posteriors of the models. Specifically, in \textit{complex subjective} tasks like multilabel emotion or morality recognition, LLMs seem to virtually disregard evidence from the dataset's mapping in their posterior predictions, even in the form of Chain-of-Thought prompting (CoT; \citealt{weiChainThoughtPrompting2022}), performing significantly worse than traditional algorithms~\cite{alhuzaliSpanemoCastingMultilabel2021, chochlakisLeveragingLabelCorrelations2023}. This \change{may imply an inability to perform} annotator modeling, as the same document can receive different but valid annotations by different people.

Here, we use \textit{complex subjective} to denote \textit{survey settings}~\cite{resnick2021survey}, in which people can reasonably disagree about their semantic interpretations, where the notion of ground truth is replaced with \textit{crowd} truth~\cite{aroyo2015truth}. 

In this work, we question whether we can efficiently use this crowd truth when modeling these tasks with LLMs. We hypothesize that the aggregation process creates artifacts in the labels provided to the model, and in turn in the prompt, causing the model to ignore the entire input-label mapping as noise. Specifically, aggregation can create \textit{inconsistent} annotations, since, for instance, different annotators can prevail in different examples, which can cause inconsistencies within or across splits \change{(for a toy example, consult Section~\ref{sec:app-inconsistency})}. We create a carefully crafted experimental setting to test our hypothesis with annotator-level labels using ICL and CoT, and try to gauge at the magnitude of this effect. We find that LLMs do indeed tend to consistently favor individual perspectives compared to the aggregate, and in fact favor minority annotators more than majority annotators, who better resemble the aggregate. Our contributions are:
\begin{itemize}
    \item  We show strong correlational evidence that aggregation creates artifacts that hinder the modeling of subjective tasks with LLMs
    \item We show that minority annotators can both align with LLMs' priors better and demonstrate larger positive effects in the posterior.
    \item We nonetheless conclude that there are more major factors hurting LLM performance in machine learning benchmarks.
\end{itemize}
We advocate for more transparency in data collection and data sharing, and urge releasing and modeling individual annotations and not simply the aggregates in benchmark datasets.

\section{Related Work}

\subsection{In-Context Learning and Priors}

ICL~\cite{brownLanguageModelsAre2020} has been used extensively to evaluate LLMs on standard benchmarks~\cite{srivastava2022beyond}. It requires no gradient-based interventions, which are otherwise costly to perform for large models, and usually achieves competitive or state-of-the-art performance. The existence of commercial APIs~\cite{achiam2023gpt}, coupled with open-weights alternatives \cite{touvronLlamaOpenFoundation2023, dubey2024llama} have made ICL an accessible generalist for language tasks and more~\cite{liu2024visual}.
Previous work has further studied controlling the reliance on context and in-weights knowledge through the distribution of the training data~\cite{chanDataDistributionalProperties2022}, examining how to optimally select examples for the prompt~\cite{rubin2022learning, gupta2023gistscore}, integrating instructions explicitly during training~\cite{touvronLlamaOpenFoundation2023, ouyang2022training}, etc. Relevant to our work, prior work has elicited ICL priors by providing random labels for the examples of the prompt~\cite{min2022rethinking}. The resulting minimal variations in performance indicates that LLMs recognize and retrieve their prior knowledge of the task in the prompt rather than doing any ``learning''.
Subsequent results challenged the view that LLMs mostly perform task recognition, showcasing a significant degradation in performance when scaling the prompt~\cite{kossen2023context} or when focusing on specific tasks instead of aggregates~\cite{yoo2022ground}, and analyzed behavior with certain label manipulations~\cite{kossen2023context, pan2023context, wei2023larger}. More recent work, however, suggests that in complex subjective tasks like emotion and morality classification, the prior understanding of the task dominates posterior predictions~\cite{chochlakis2024strong}.

One potential way to augment ICL and overcome the prior bias is with CoT~\cite{weiChainThoughtPrompting2022}. CoT incorporates the derivation process from input to output explicitly in the prompt, presenting a more human-like reasoning process. This has several advantages, such as making some patterns in the data explicit in the prompt, making model responses more explainable, and potentially directing more computing resources towards more complex problems. However, detailed analysis has cast some doubt on the reliability and the faithfulness of this reasoning technique~\cite{lanham2023measuring, turpin2024language}. Nonetheless, CoT seems to improve the robustness and performance of LLMs across a plethora of tasks, yet follow-up work on subjective tasks showed no improvements and the same bias towards a \textit{reasoning prior}, especially for larger models~\cite{chochlakis2024larger}. Methods such as Tree of Thoughts~\cite{yao2024tree} or self-consistency~\cite{wang2022self} have experimented with ways to further augment CoT. In this work, we study whether part of the prior bias in both ICL and CoT can be explained by the aggregation process used to derive the labels, and in turn the reasoning process, of the examined benchmarks.

\subsection{Annotator Disagreement and Modeling}

Many works have attempted to model individual annotator perspectives \change{instead of the aggregate, like we advocate in this work. For example}, researchers used the EM algorithm~\cite{dawidMaximumLikelihoodEstimation1979a} to assign confidence to each annotator's evaluations~\cite{hovy2013learning}. Recently, \citet{gordonJuryLearningIntegrating2022} concatenated features derived from Transformers~\cite{vaswaniAttentionAllYou2017} with annotator embeddings that incorporate demographics to model individual perspectives. Demographic information has also been incorporated in word embeddings by ~\citet{gartenIncorporatingDemographicEmbeddings2019}. Demographic information and psychological profiles have been statistically examined in text annotations to derive insights into systematic biases~\cite{sapAnnotatorsAttitudesHow2022}. Recent work has tried to filter annotators based on deep learning methods~\cite{mokhberian2022noise}, to model annotators on top of common representations~\cite{davaniDealingDisagreementsLooking2022, mokhberian2023capturing}, and to decrease annotation costs based on agreement~\cite{golazizian2024cost}. Modeling annotators with LLMs has shown limited success, and LLM biases have also been explored~\cite{dutta2023modeling, abdurahman2024perils, hartmann2023political}.

\section{Methodology}

We closely follow the methodology and notation of \citet{chochlakis2024larger}. For a set of examples $\mathcal{X}$, and a set of labels $\mathcal{Y}$, a dataset $\mathcal{D}^a$ defines a mapping $f^a:\mathcal{X} \rightarrow \mathcal{Y}$, where $a$ denotes a specific annotator or the aggregate, as well as reasoning chains $R^a(x) = r, x\in\mathcal{X}$ that explicitly describe $f^a$, and therefore \mbox{$\mathcal{D}^a = \{(x, y, r): x \in \mathcal{X}, y = f^a(x), r = R^a(x)\}$}, from which we can sample demonstrations with $p(x, y, r)$. We do not differentiate between splits for brevity. Given CoT prompt \mbox{$S = \{(x_i, y_i, r_i): (x_i, y_i, r_i) \sim p_S, i \in [k]\}$} with $k$ demonstrations sampled with distribution $p_S$ from $\mathcal{D}^a$ without replacement, an LLM produces its own mapping and predictions for the task, denoted as \mbox{$\hat{f}_k(.; p_S): \mathcal{X} \rightarrow \mathcal{Y}$}. When using regular ICL, we simply drop the reasoning text. For all our experiments, we set the temperature $\tau = 0$ to derive deterministic predictions, and vary the examples and/or labels between runs.

\subsection{Similarity and Performance Metrics} \label{sec:metrics-def}

To keep evaluations consistent when using API-based LLMs, we rely on similarity measures calculated directly on the final predictions rather than probabilistic measures like logits. Therefore, we use the Jaccard Score (JS), Micro and Macro F1 metrics~\cite{mohammad2018semeval} to evaluate the performance of the models. For consistency, we also use them to quantify the similarity between the predictions from different model runs or annotators, since they are symmetric functions which we apply to interchangeable sets.

\begin{figure}[t]
    \includegraphics[width=1\columnwidth]{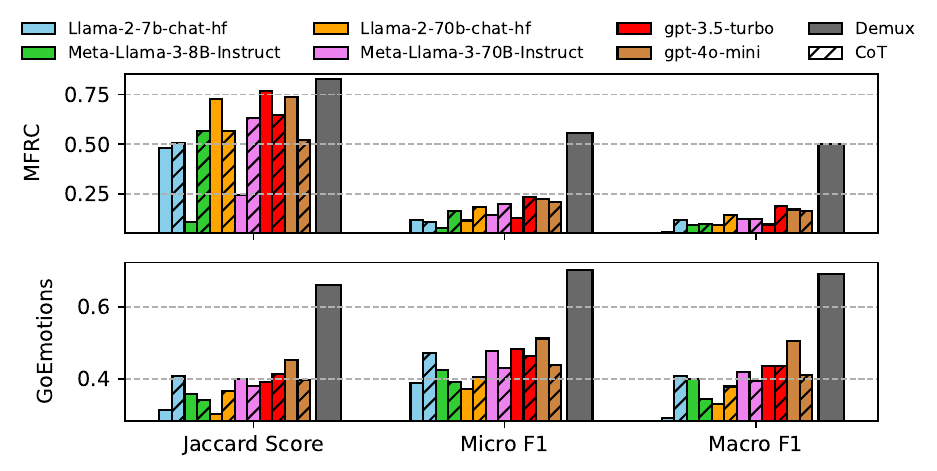}
    \caption{Performance comparison between LLMs w/ and w/o Chain-of-Thought prompting compared to Demux (BERT-based) using aggregated labels.}
    \label{fig:baselines}
\end{figure}

\begin{figure*}[t]
    \centering
  \includegraphics[width=0.95\linewidth]{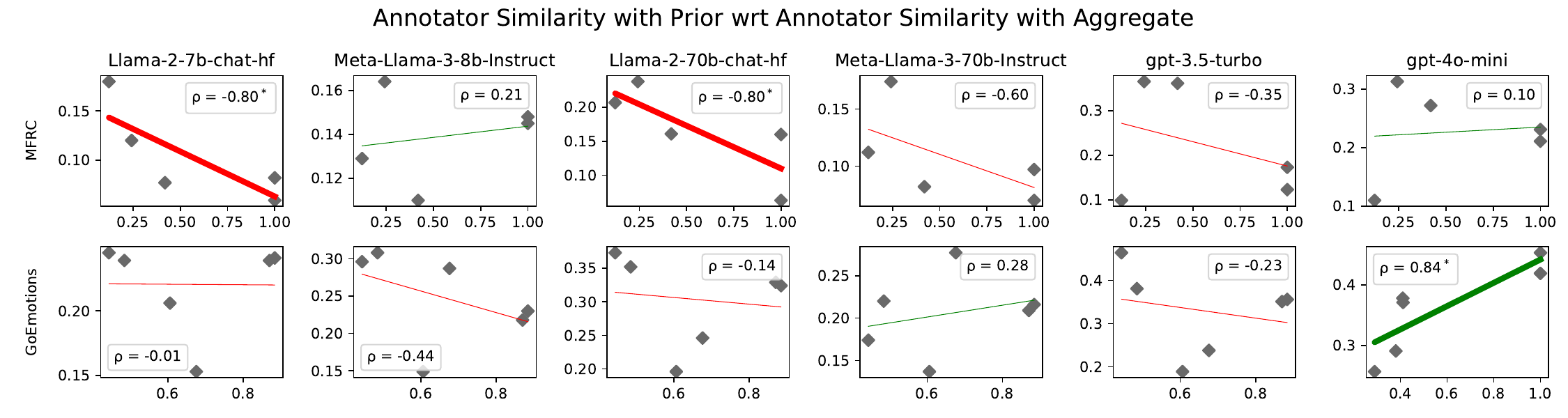}
  \caption{Scatter plot of annotator similarity with aggregate and with prior. \change{Correlation and line fit of data also shown. $^*$ and bold lines: $p < 0.05$.}}
  \label{fig:prior-aggr}
\end{figure*}

\subsection{Task Prior Knowledge Proxies via Zero-shot Inference}

Here, we precisely define the priors that we use for ICL and CoT. \change{We use the term \textbf{prior} to contrast it with the \textbf{posterior} predictions of the model after \textit{evidence} (i.e., a specific annotator's labels) from the dataset have been presented to it}. First, we have the true \textbf{reasoning task-recognition zero-shot\footnote{since no annotations are required} prior}, where the prompt contains $k$ demonstrations sampled with \mbox{$p^I(x, y, r) = p(x) p(y) p(r)$}, so text, labels, and reasoning are sampled \textit{independently} from each other from $\mathcal{D}^a$, hence labels and reasoning are irrelevant to the text and each other. This effectively maintains the relationships between labels, which are strong in such multilabel tasks~\cite{cowenSelfreportCaptures272017}. For ICL, we have the corresponding \textbf{task-recognition zero-shot prior} sampled with \mbox{$p^I(x, y) = p(x) p(y)$}, so text and labels are also sampled independently. The similarity of the priors to annotators and aggregate are, therefore, measured by comparing (as described in Section~\ref{sec:metrics-def}) the prior predictions $\hat{f}_k(.; p_I)$ with the annotator's labels, which is equivalent to the prior performance for the annotator, and the posteriors are simply defined as the $\hat{f}_k(.; p_S)$ with the joint sampling distribution $p_S$ (meaning we present the gold labels to the model). 

\subsection{Prompt Design}

Because the specific examples and their order in the prompt can affect the output of the model, we use exactly the same examples and in the same order across corresponding experiments. To achieve that, we find groups of annotators with significant overlap in the train and in the evaluation sets and use the same samples, including for aggregate and prior. Since the only degree of freedom is the labels, we eliminate \change{the aforementioned} confounding factors, \change{among others} (more details in Section~\ref{sec:prompt}).

\section{Experiments}

\begin{figure*}[t]
    \centering
  \includegraphics[width=0.98\linewidth]{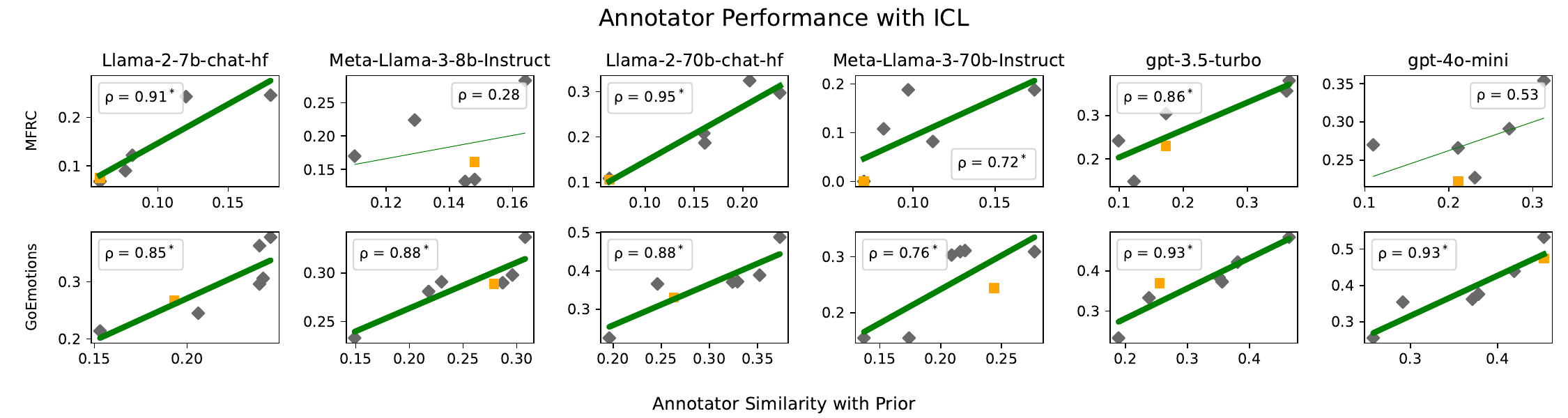}
  \includegraphics[width=0.98\linewidth]{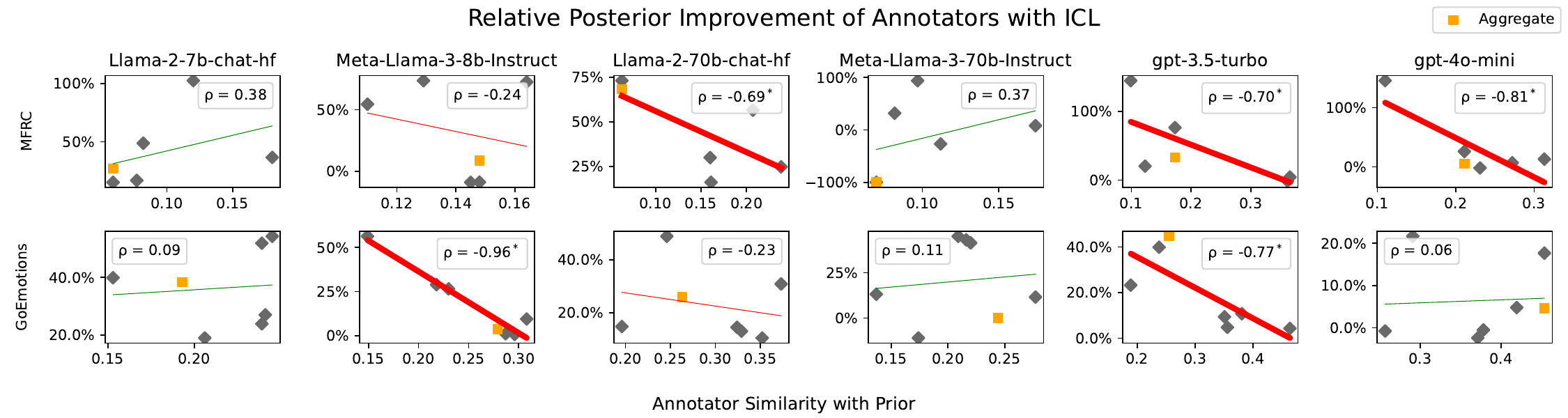}
  \caption {\textbf{In-Context Learning} performance for annotators and aggregate and their relative improvement compared to the model's prior as a function of the similarity of each with the model's prior. \change{Correlation and line fit of data also shown. $^*$ and bold lines: $p < 0.05$.}}
  \label{fig:prior}
\end{figure*}

\begin{figure*}[!]
  \centering
  \includegraphics[width=0.98\linewidth]{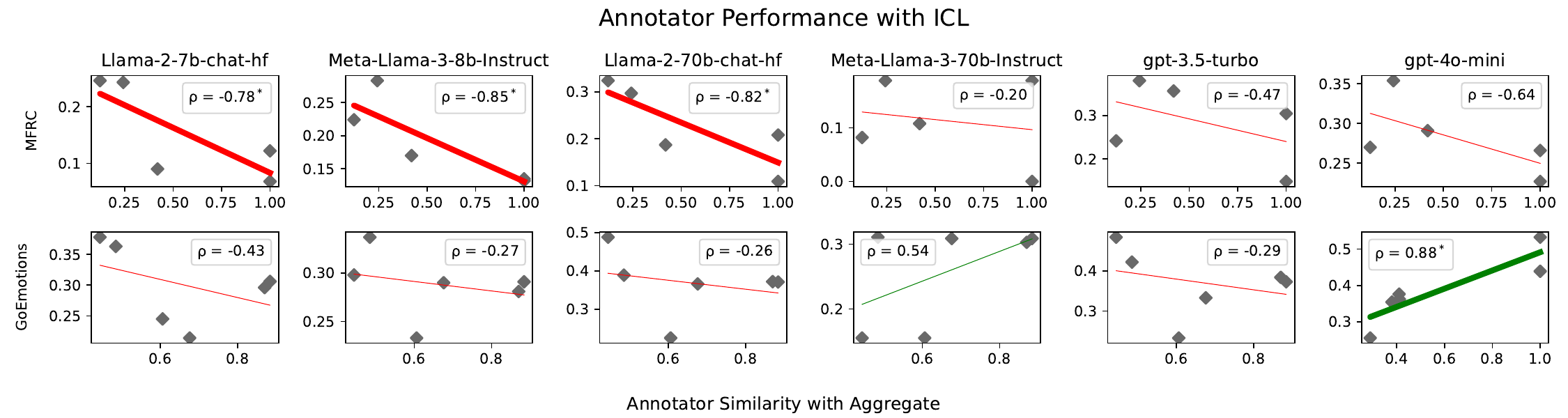}
  \includegraphics[width=0.98\linewidth]{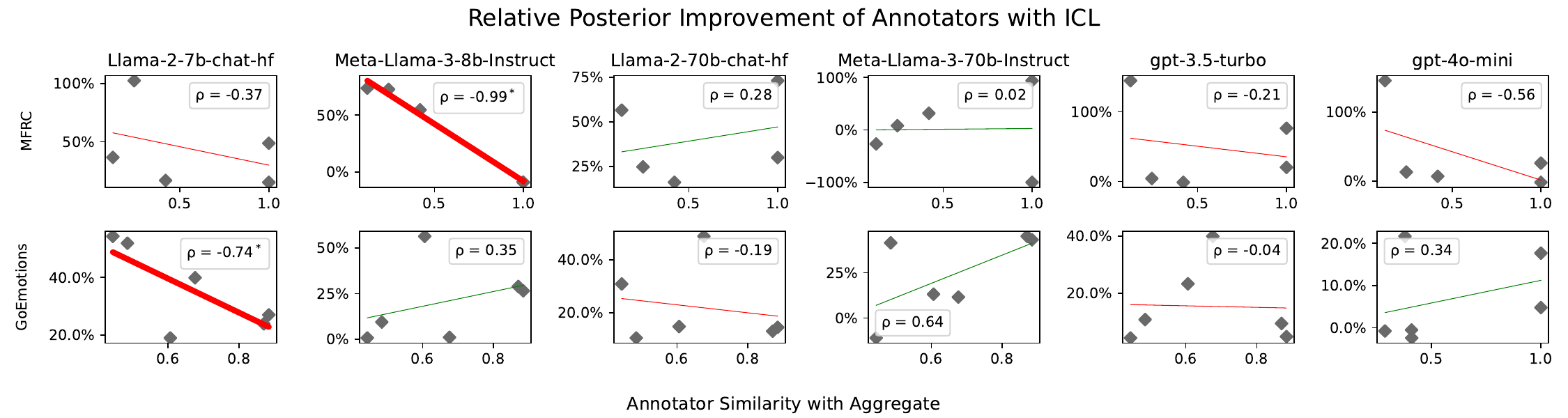}
  \caption {\textbf{In-Context Learning} performance for annotators and their relative improvement compared to the model's prior as a function of the similarity of each annotator with the aggregate. \change{Correlation and line fit of data also shown. $^*$ and bold lines: $p < 0.05$.}}
  \label{fig:aggr}
\end{figure*}

\begin{figure*}[t]
  \centering
  \includegraphics[width=0.98\linewidth]{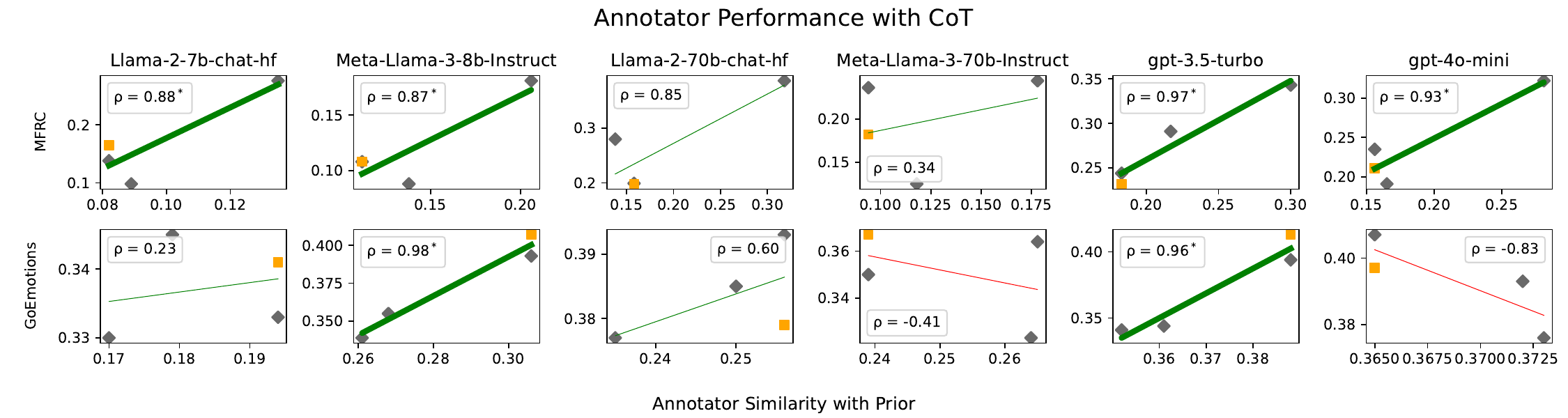}
  \includegraphics[width=0.98\linewidth]{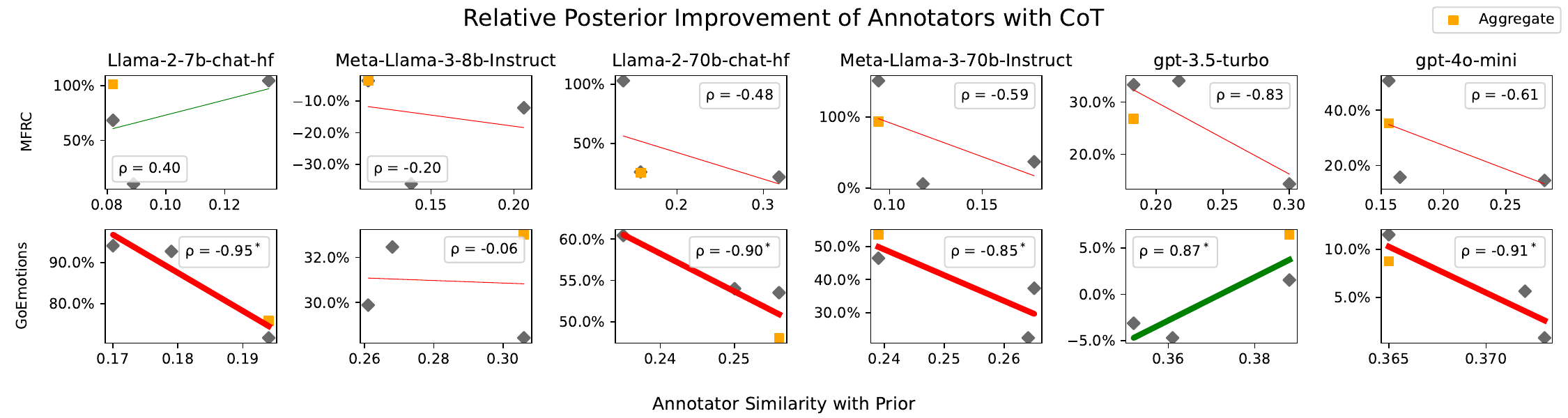}
  \caption {\textbf{Chain-of-Thought} performance for annotators and aggregate and their relative improvement compared to the model's prior as a function of the similarity of each with the model's prior. \change{Correlation and line fit of data also shown. $^*$ and bold lines: $p < 0.05$.}}
  \label{fig:prior-cot}
\end{figure*}

\begin{figure*}[t]
  \includegraphics[width=0.98\linewidth]{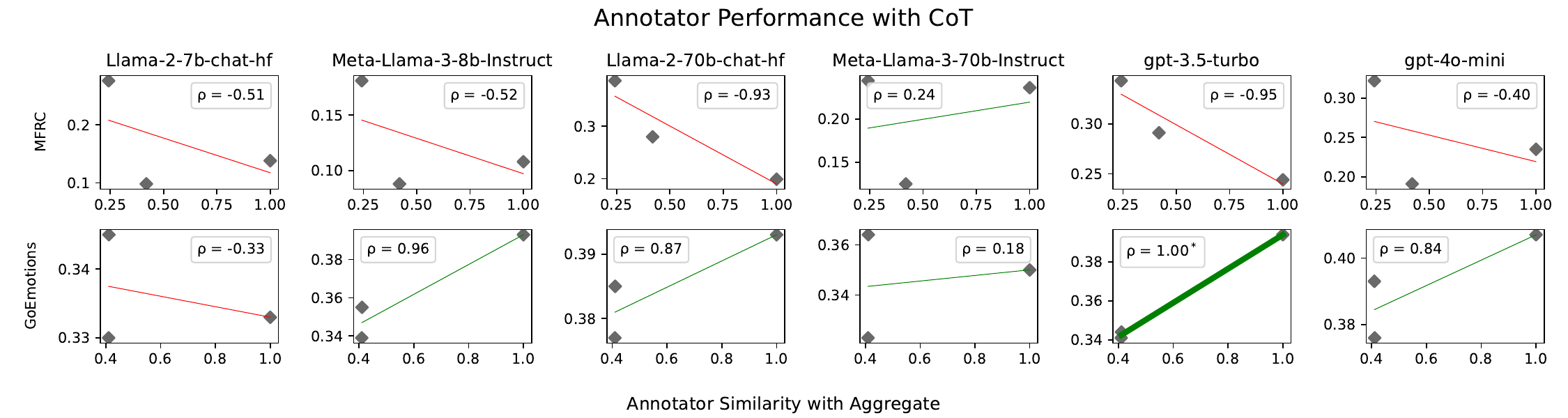}
  \includegraphics[width=0.98\linewidth]{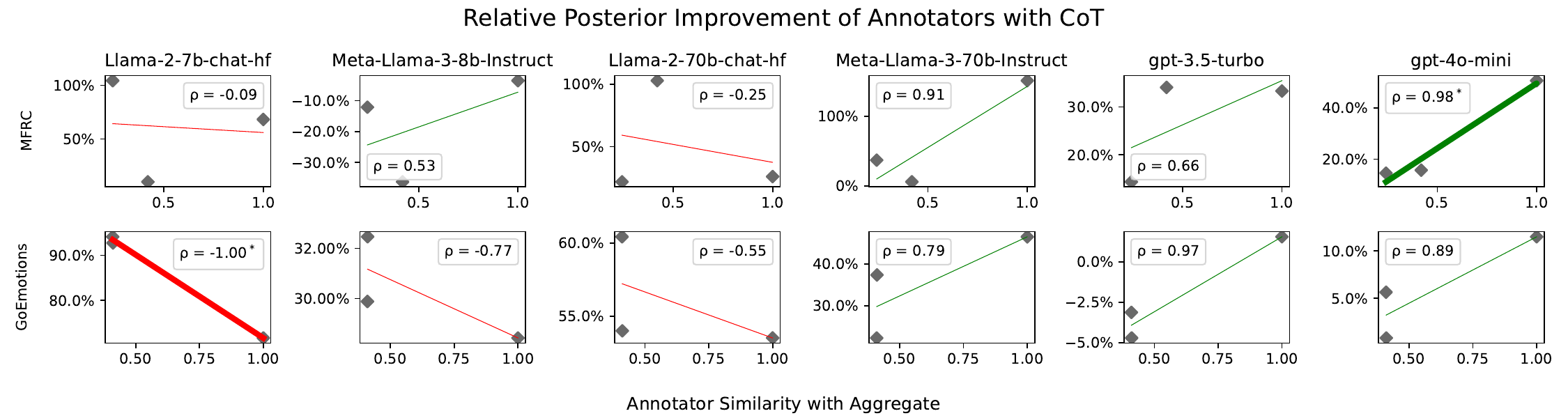}
  \caption {\textbf{Chain-of-Thought} performance for annotators and their relative improvement compared to the model's prior as a function of the similarity of each annotator with the aggregate. \change{Correlation and line fit of data also shown. $^*$ and bold lines: $p < 0.05$.}}
  \label{fig:aggr-cot}
\end{figure*}

\subsection{Datasets}

\noindent \textbf{MFRC}~\cite{trager2022moral}: Multilabel moral foundation corpus with annotations for six moral foundations: \textit{care}, \textit{equality}, \textit{proportionality}, \textit{loyalty}, \textit{authority}, and \textit{purity}. We use annotators \texttt{00} through \texttt{04} (common examples between groups \texttt{00}-\texttt{01}-\texttt{03} and \texttt{02}-\texttt{04}).

\noindent \textbf{GoEmotions} \cite{demszky2020goemotions}: Multilabel emotion recognition benchmark with 27 emotions. For efficiency and conciseness, we pool the emotions to the following seven ``clusters'' by using hierarchical clustering: \textit{admiration}, \textit{anger}, \textit{fear}, \textit{joy}, \textit{optimism}, \textit{sadness}, and \textit{surprise}. We use annotator triplets \texttt{4}-\texttt{37}-\texttt{61} and \texttt{7}-\texttt{36}-\texttt{60}.

\subsection{Implementation Details}

We use the 4-bit quantized versions of the open-source LLMs through the \textit{HuggingFace}~\cite{wolf-etal-2020-transformers} interface for \textit{PyTorch}. We use LLaMA-2 7B and 70B, LLaMA-3 8B and 70B, GPT-3.5 Turbo, and GPT-4o mini. We chose only models with RLHF~\cite{ouyang2022training} \change{for uniformity}. We perform 3 runs for each LLM experiment, varying the examples used. Statistical significance is calculated with permutation tests and measured by considering all 3 runs as separate data points. We use random retrieval of examples. We use less shots for CoT given the increases in length in the prompt. We generated reasonings for each example per annotator and for the aggregate. \change{For details on our CoT annotations, see Section~\ref{sec:annotations} in the Appendix}. We use one NVIDIA A100 and one V100.

\subsection{Baselines}

To establish baseline performance of LLMs compared to smaller, gradient-based methods, we present performance with and without CoT prompting compared to BERT-based~\cite{devlin2018bert} Demux~\cite{chochlakisLeveragingLabelCorrelations2023}. In Figure~\ref{fig:baselines}, we demonstrate the significant difference in performance across all the LLMs (45 shots for ICL, 15 shot for CoT) and Demux. In fact, given the very high JS and very low F1 scores, results for MFRC indicate close to random performance for the model, \change{so we choose to use Micro F1 for MFRC, as opposed to JS for GoEmotions}.

Nevertheless, we argue that this is an artifact of the inconsistent mapping used in the prompt, caused by the aggregation of labels for different annotators. Next, we evaluate whether aggregation does create annotation artifacts, and the extend to which they influence model behavior.

\subsection{Main Results}

In this section, we present our experiments, aimed at disentangling the role of aggregation in subjective tasks. First, we focus on 45-shot ICL (Section~\ref{sec:icl}), and analyze how the similarity of each annotator and of the aggregate with the models' prior affects the relative improvement of the posterior $p_S$ over the prior $p_I$, as well as absolute posterior performance. Then, we analyze how the majority and minority (or idiosyncratic) annotators fare on these tasks by looking at the similarity of each annotator to the aggregate. Following this analysis, we perform equivalent experiments for 15-shot CoT, and present them in Section~\ref{sec:cot}.
% TODO: added a couple of sections in between, talk about these
Finally, we zoom out of individual experiments and summarize our complete body of evidence in Section~\ref{sec:conclusion}.

In all experiments, we present averages. The similarity of the annotators to the prior, as well as the similarity to the aggregate, is measured on the test set, which may differ from the train set.

First of all, we present the relationship between the similarity of each annotator to the aggregate and the similarity to the prior of the model. Our results in Figure~\ref{fig:prior-aggr} indicate that the correlation tends to be negative, with 7 of 12 negative correlations, 2 of which are statistically significant, and only 4 positive correlations with one statistically significant result. This indicates that the more an annotator resembles the majority, the less aligned they tend to be with the models' priors. This is the first piece of evidence indicating that aggregation causes artifacts, and seems to suggest that the models perform more appropriate aggregation of evidence during their training compared to the simple (or even simplistic) majority-based aggregation used in such benchmarks, causing misalignment.

\subsubsection{Annotator Modeling with ICL} \label{sec:icl}

\paragraph{Similarity to Prior} 

In Figure~\ref{fig:prior}, we first see the performance of the models for each annotator and the aggregate w.r.t. the similarity of each to the prior of each model. For the performance of the model, we see that, as expected, similarity with the prior correlates positively with final performance, with all results but one being statistically significant. It is interesting to see that the aggregate ranks low both for MFRC and GoEmotions in terms of posterior (from left to right and top to bottom: 5/6, 4/6, 6/6, 6/6, 5/6, 6/6, 5/7, 5/7, 6/7, 5/7, 5/7, 2/7; average is 22nd percentile) and prior (5/6, 2/6, 6/6, 6/6, 4/6, 5/6, 6/7, 4/7, 5/7, 2/7, 5/7, 2/7; average is 33rd percentile). Looking at the relative improvement, it is interesting to see that the only significant trends are negative trends, meaning that the LLMs tend to boost opinions they disagree with more. Despite the aggregate being among the worst performing mappings, with the expectation being that it receives significant gains in performance, it ranked below average (4/6, 4/6, 2/6, 6/6, 3/6, 5/6, 4/7, 5/7, 3/7, 6/7, 1/7, 4/7; average is 39th percentile).

\paragraph{Agreement with Aggregate} By switching to examining at the similarity of each annotator to the aggregate, and how that correlates with absolute and relative performance, in Figure~\ref{fig:aggr}, we see strongly negative trends. In fact, 17 out of the 24 cases are negative, 6 of which are statistically significant, and only one positive trend is statistically significant. Therefore, we see that idiosyncratic annotators both have better performance and are more amplified.

Overall, we see very strong correlational evidence that not only are aggregates misaligned with the models' priors, they also benefit less from ICL with their labels in the prompt. This is happening in spite of annotators with worse alignment frequently receiving significant gains in performance. Consequently, by combining our findings from the priors and the aggregates, we conclude that the aggregate mapping inherently has inconsistencies that inject sufficient noise in the prompt.

\subsubsection{Annotator Modeling with CoT} \label{sec:cot}

We now switch to CoT, and evaluate consistency across prompting techniques. We note that because of the decreased number of runs in this setting, the confidence in these findings is similarly decreased.

\paragraph{Similarity to Prior} In Figure~\ref{fig:prior-cot}, we observe similar trends to the equivalent setting for ICL (Figure~\ref{fig:prior}), namely that final (posterior) performance positively correlates with the prior similarity (10 of 12 cases are positive, 6 of which are statistically significant), and that relative improvement of the posterior compared to the prior is negatively correlated with similarity with the prior (10 of 12 cases are negative, 4 of which are statistically significant). That being said, the differences in performance in Figure~\ref{fig:prior-cot} tend to be smaller than in Figure~\ref{fig:prior}. The aggregate is among the worst performers in MFRC, but the results in GoEmotions are equivocal.

\paragraph{Agreement with Aggregate} In Figure~\ref{fig:aggr-cot}, we present CoT results and correlate them with the similarity to the aggregate. Our experiments here seem to be split between negative and positive trends.

Here, due to the decreased number of experiments, it is difficult to extract concrete findings, yet we can ascertain that the findings here do not seem to contradict our previous findings, and do no show improved performance compared to ICL modeling.

\subsection{Detailed Analysis} \label{sec:details}

\begin{table*}[]
    \centering
    \begin{tabular}{lcccc}
        \multirow{3}{*}{\textbf{Model}} & \multicolumn{2}{c}{Authority \textbf{F1}} & \multicolumn{2}{c}{Equality \textbf{F1}}  \\
        \cmidrule(lr){2-3} \cmidrule(lr){4-5}
        & Prior & Post & Prior & Post \\
        \midrule
        \texttt{Llama-2-7b-chat-hf} ICL & \rescell{0.135}{0.121} & \textcolor{green}{$\uparrow$} \rescell{0.217}{0.153} & \rescell{0.204}{0.059} & \textcolor{red}{$\downarrow$} \rescell{0.133}{0.189} \\
        \texttt{Llama-2-7b-chat-hf} CoT & \rescell{0.157}{0.124} & \textcolor{green}{$\Uparrow$} \rescell{0.330}{0.030} & \rescell{0.524}{0.140} & \textcolor{red}{$\Downarrow$} \rescell{0.197}{0.004} \\
        \texttt{Meta-Llama-3-8B-Instruct} ICL & \rescell{0.115}{0.083} & \textcolor{green}{$\Uparrow$} \rescell{0.276}{0.033} & \rescell{0.039}{0.055} & \textcolor{green}{$\uparrow$} \rescell{0.074}{0.105} \\
        \texttt{Meta-Llama-3-8B-Instruct} CoT & \rescell{0.082}{0.116} & \textcolor{green}{$\Uparrow$} \rescell{0.277}{0.044} & \rescell{0.300}{0.216} & \textcolor{green}{$\uparrow$} \rescell{0.481}{0.086}  \\
        \texttt{Llama-2-70b-chat-hf} ICL & \rescell{0.035}{0.050} & \textcolor{green}{$\uparrow$} \rescell{0.053}{0.075} & \rescell{0.080}{0.063} & \textcolor{green}{$\uparrow$} \rescell{0.133}{0.189} \\
        \texttt{Llama-2-70b-chat-hf} CoT & \rescell{0.107}{0.097} & \textcolor{green}{$\Uparrow$} \rescell{0.257}{0.053} & \rescell{0.044}{0.063} & \textcolor{green}{$\Uparrow$} \rescell{0.250}{0.041}  \\
        \texttt{Meta-Llama-3-70B-Instruct} ICL & \rescell{0.263}{0.118} & \textcolor{green}{$\Uparrow$} \rescell{0.347}{0.063} & \rescell{0.129}{0.118} & \textcolor{green}{$\uparrow$} \rescell{0.269}{0.234} \\
        \texttt{Meta-Llama-3-70B-Instruct} CoT & \rescell{0.242}{0.121} & \textcolor{green}{$\uparrow$} \rescell{0.383}{0.047} & \rescell{0.243}{0.096} & \textcolor{green}{$\uparrow$} \rescell{0.381}{0.135} \\
        \texttt{gpt-3.5-turbo} ICL & \rescell{0.319}{0.057} & \textcolor{green}{$\uparrow$} \rescell{0.357}{0.066} & \rescell{0.167}{0.236} & \textcolor{green}{$\Uparrow$} \rescell{0.468}{0.100} \\
        \texttt{gpt-3.5-turbo} CoT & \rescell{0.253}{0.068} & \textcolor{green}{$\Uparrow$} \rescell{0.361}{0.057} & \rescell{0.260}{0.131} & \textcolor{green}{$\uparrow$} \rescell{0.350}{0.048} \\
        \texttt{gpt-4o-mini} ICL & \rescell{0.228}{0.188} & \textcolor{green}{$\Uparrow$} \rescell{0.399}{0.014} & \rescell{0.394}{0.193} & \textcolor{red}{$\downarrow$} \rescell{0.245}{0.204} \\
        \texttt{gpt-4o-mini} CoT & \rescell{0.296}{0.039} & \textcolor{red}{$\downarrow$} \rescell{0.287}{0.053} & \rescell{0.166}{0.060} & \textcolor{green}{$\uparrow$} \rescell{0.222}{0.057} \\
    
    \end{tabular}
    \caption{Comparison of prior and posterior F1 score of authority and equality for annotator \texttt{01} of MFRC using ICL or CoT. $\uparrow/\downarrow$ : increase/decrease with overlapping ranges, $\Uparrow/\Downarrow$ : increase/decrease with non-overlapping ranges.}
    \label{tab:ann01}
\end{table*}

Based on observations during our manual annotation efforts, we identified clear patterns that annotator \texttt{01} in MFRC provides the \textit{Authority} label more frequently even when one authority figure is mentioned in the input, in contrast to the rest of the annotators. Therefore, if any learning is achieved from ICL and CoT, we expect the accuracy for \textit{Authority} to be visibly improved compared to the prior baseline due to the consistency and the clear pattern. To achieve and test that, we made this implicit bias clear in the generated reasoning chains. We present the change in performance in Table~\ref{tab:ann01} in comparison to another label, \textit{Equality}, chosen at random, since we did not observe any other clear patterns in other labels. We do indeed observe that the gains in \textit{Authority} F1 score are consistent and tend to be significant across models and settings (with only GPT-4o mini CoT presenting an insignificant drop), in opposition to \textit{Equality}, where gains tend to be small and insignificant, and large drops in performance are observed. This does indicate an ability for the models to somewhat learn and revise priors from the prompt \textit{when the mapping and/or the rationale presented are consistent and clear}.

\subsection{Best Performing Annotators} 
% [{'annotator01': 'skyblue', 'annotator00': 'limegreen', 'annotator03': 'dimgray', 'annotator04': 'violet', 'annotator02': 'red', 'aggr': 'orange'}, {'4': 'skyblue', '61': 'limegreen', '37': 'dimgray', '60': 'violet', '7': 'red', '36': 'peru', 'aggr': 'orange'}]
% MFRC: annotator01 gpt3.5 0.35
% GoEmotions: 60 gpt4o 0.52

\begin{figure}[t]
    \hspace{-10px}
    \includegraphics[width=1.07\columnwidth]{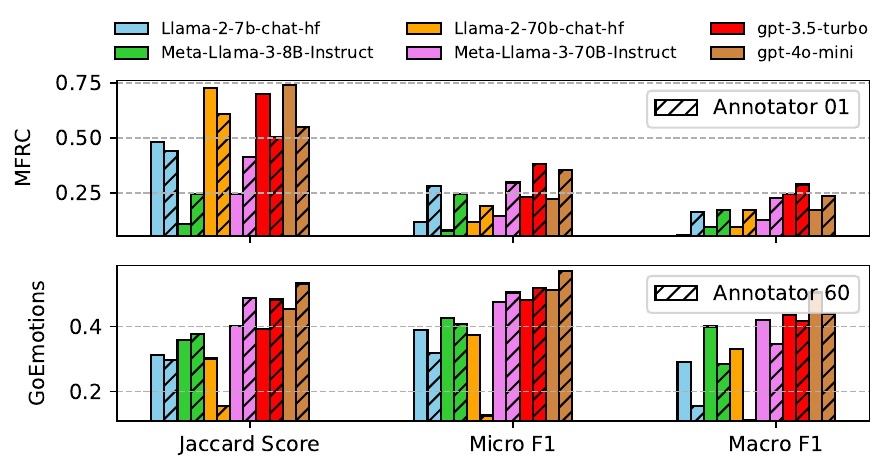}
    \caption{Performance comparison across LLMs using ICL when using aggregated labels and using best-case scenario individual annotator labels.}
    \label{fig:annotators}
\end{figure}

Finally, and as a best case scenario, we present the performance of the best annotator with ICL (based on performance across models as can be seen in Figures~\ref{fig:prior} and \ref{fig:aggr}) and compare that to the aggregate. Results are shown in Figure~\ref{fig:annotators}. The benefits from modeling a specific annotator are evident, as we observe large gains in performance in two out of the three metrics, a finding that is consistent across all models except for LLaMA-2 70b. This further emphasizes that modeling individual perspectives instead of aggregates is beneficial.

\section{Conclusion} \label{sec:conclusion}

In summary, we see that aggregated labels tend to align less with the LLMs' prior for each task. Furthermore, and in spite of worse aligned annotators receiving larger posterior performance increases, the aggregate posterior appears to collapse to the prior, resulting in significantly worse performance to several annotators. This result indicates that the majority is not necessarily well-aligned with models. Then, it is evident that interpretable and consistent mappings can be modeled by LLMs and improve upon the prior, even though the model might not align with the specific annotator a priori. Finally, we see that individuals do indeed result in better performance on the task.

Given the commonsense reasoning capabilities of LLMs, the emotional and moral capabilities of LLMs demonstrated in settings different from traditional machine learning settings~\cite{tak2024gpt}, as well as best-case results, we conclude that the \textbf{aggregation process introduces artifacts} in the labels that cause LLMs to ignore the mapping as noise. It is interesting to note that in our prompts, the aggregate in rarely a hodgepodge of disparate opinions, wherein the aggregate does not match any individual, but factors like different annotator mixtures, especially between train and test splits, as well as different annotator groups prevailing in different examples, introduce sufficient noise in the annotations that cannot be modeled with ICL and CoT. That being said, the performance gap with gradient-based methods remains large, suggesting that other factors like task complexity also majorly account for the observed biasing effects.

Finally, we question what it means to model these aggregate opinions. Namely, since they might not reflect the opinion of any individual, even if in a minority of cases, then what rationale should be provided and how should it be generated in a sound manner? We advocate, therefore, as previous work has done~\cite{prabhakaranReleasingAnnotatorlevelLabels2021, dutta2023modeling}, for releasing and modeling annotator-level labels instead of aggregates, and suggest that the field of \textbf{subjective modeling should move away from aggregate modeling} in the age of LLMs and of more elaborate modeling methods such as CoT.

\section{Limitations}

Given our constraints to standardize the prompt and remove other degrees of freedom that can constitute potential confounding factors in our evaluations, but also for computational efficiency, the evaluations sets contain a small number of examples (namely, 100 and 71 for the triplets in GoEmotions, and 100 for both groups in MFRC), increasing the noise in our findings. Nonetheless, this practice has become standard in the evaluation of LLMs.

\change{A potential confounding factor that we do not control for is the quantization, as previous work has reported significant decreases in performance from it~\cite{marchisio2024does}. We note, first, that there is no reason for the quantization to affect our results in a nonuniform way, second, that we perform the quantization because of obvious computational constraints, and third, it is possible that even some API-based models are served quantized (e.g., Turbo versions). For these reasons, we believe that quantized performance is representative of LLM performance in realistic scenarios. Moreover, this work does not aim to establish the performance of LLMs in these subjective tasks compared to other models, but rather to compare within the LLMs themselves. Nonetheless, we perform experiments with various levels of quantization of LLaMA-3 8B, and show the results in Table~\ref{tab:quant} in the Appendix.}

In our study of specific labels and the effects of what we perceive as consistency on the performance of LLMs (Section~\ref{sec:details}), a potential confounding factor in analysis is the increased frequency of the label, since the studied annotator was more sensitive to specific stimuli in the input, as described. We specifically chose to perform a more detailed analysis in this positive pattern, however, because we expect the models, as did we, find it easier to distinguish positive patterns in the data.

It is important to note that we have performed experiments in only two problems and benchmarks, as we opted in favor of presenting more LLMs. Therefore, our findings may not generalize to other, highly subjective tasks. Furthermore, other datasets with stricter annotation manuals that aim to resolve all ambiguities may not present similar behavior, as annotator agreement is artificially raised by removing some of the subjectivity of the semantic interpretations of the annotators in favor of following stricter, highly specific instructions.

We also want to note that we do not perform Bonferonni correction across models and datasets given the small number of datapoints we have to compute our correlations, yet we believe it is important to highlight the settings with smaller p-values.

Moreover, datasets with data derived from social media have been criticized as lacking the context for a model---or even humans---to make appropriate judgments about the emotion or morality expressed in them \cite{yang2023context}, and techniques to evaluate the correctness of the labels in a dataset have been designed to discard noisy samples \cite{swayamdipta2020dataset, mokhberian2022noise}. Given the subjective nature of the task and the lack of context, such tools could be used to perhaps improve performance. That being said, the interpretation by humans is sufficiently consistent for Demux to achieve better performance than every LLM. We also note that by removing ambiguous examples, we may also remove that which makes these tasks challenging \cite{aroyo2015truth}.

Finally, while we carefully control the experimental setting to control for confounding factors, we do not actually perform causal interventions, and consequently present correlational evidence.

\section{Ethical Considerations}

Our focus on traditional machine learning benchmarks, as well as our takeaways, should complement and not compete with quantifying bias in LLMs using other tools and techniques \cite{caliskan2017semantics, gonen2019lipstick, ferrara2023should, abdurahman2024perils}. It is also important to emphasize that improving affective and moral capabilities in LLMs entails perils, since better catering the emotional and moral responses to more contexts and personalizing them to specific individuals can lead to improved manipulation of users by LLMs.

\vspace{-5px}
\section*{Acknowledgments}

This project was funded in part by DARPA under contract HR001121C0168, and in part by NSF CIVIC.

\bibliography{library}

\begin{thebibliography}{56}
\providecommand{\natexlab}[1]{#1}

\bibitem[{Abdurahman et~al.(2024)Abdurahman, Atari, Karimi-Malekabadi, Xue, Trager, Park, Golazizian, Omrani, and Dehghani}]{abdurahman2024perils}
Suhaib Abdurahman, Mohammad Atari, Farzan Karimi-Malekabadi, Mona~J Xue, Jackson Trager, Peter~S Park, Preni Golazizian, Ali Omrani, and Morteza Dehghani. 2024.
\newblock Perils and opportunities in using large language models in psychological research.
\newblock 3(7):pgae245.

\bibitem[{Achiam et~al.(2023)Achiam, Adler, Agarwal, Ahmad, Akkaya, Aleman, Almeida, Altenschmidt, Altman, Anadkat et~al.}]{achiam2023gpt}
Josh Achiam, Steven Adler, Sandhini Agarwal, Lama Ahmad, Ilge Akkaya, Florencia~Leoni Aleman, Diogo Almeida, Janko Altenschmidt, Sam Altman, Shyamal Anadkat, et~al. 2023.
\newblock \href {https://arxiv.org/abs/2303.08774} {Gpt-4 technical report}.

\bibitem[{Alhuzali and Ananiadou(2021)}]{alhuzaliSpanemoCastingMultilabel2021}
Hassan Alhuzali and Sophia Ananiadou. 2021.
\newblock \href {https://arxiv.org/abs/2101.10038} {Spanemo: {{Casting}} multi-label emotion classification as span-prediction}.

\bibitem[{Aroyo and Welty(2015)}]{aroyo2015truth}
Lora Aroyo and Chris Welty. 2015.
\newblock Truth is a lie: {{Crowd}} truth and the seven myths of human annotation.
\newblock 36(1):15--24.

\bibitem[{Brown et~al.(2020)Brown, Mann, Ryder, Subbiah, Kaplan, Dhariwal, Neelakantan, Shyam, Sastry, and Askell}]{brownLanguageModelsAre2020}
Tom Brown, Benjamin Mann, Nick Ryder, Melanie Subbiah, Jared~D Kaplan, Prafulla Dhariwal, Arvind Neelakantan, Pranav Shyam, Girish Sastry, and Amanda Askell. 2020.
\newblock Language models are few-shot learners.
\newblock 33:1877--1901.

\bibitem[{Caliskan et~al.(2017)Caliskan, Bryson, and Narayanan}]{caliskan2017semantics}
Aylin Caliskan, Joanna~J Bryson, and Arvind Narayanan. 2017.
\newblock Semantics derived automatically from language corpora contain human-like biases.
\newblock \emph{Science}, 356(6334):183--186.

\bibitem[{Chan et~al.(2022)Chan, Santoro, Lampinen, Wang, Singh, Richemond, McClelland, and Hill}]{chanDataDistributionalProperties2022}
Stephanie~CY Chan, Adam Santoro, Andrew~Kyle Lampinen, Jane~X Wang, Aaditya~K Singh, Pierre~Harvey Richemond, James McClelland, and Felix Hill. 2022.
\newblock Data distributional properties drive emergent in-context learning in transformers.
\newblock In \emph{Advances in {{Neural Information Processing Systems}}}.

\bibitem[{Chochlakis et~al.(2023)Chochlakis, Mahajan, Baruah, Burghardt, Lerman, and Narayanan}]{chochlakisLeveragingLabelCorrelations2023}
Georgios Chochlakis, Gireesh Mahajan, Sabyasachee Baruah, Keith Burghardt, Kristina Lerman, and Shrikanth Narayanan. 2023.
\newblock Leveraging label correlations in a multi-label setting: {{A}} case study in emotion.
\newblock In \emph{{{ICASSP}} 2023-2023 {{IEEE International Conference}} on {{Acoustics}}, {{Speech}} and {{Signal Processing}} ({{ICASSP}})}, pages 1--5. IEEE.

\bibitem[{Chochlakis et~al.(2025)Chochlakis, Pandiyan, Lerman, and Narayanan}]{chochlakis2024larger}
Georgios Chochlakis, Niyantha~Maruthu Pandiyan, Kristina Lerman, and Shrikanth Narayanan. 2025.
\newblock Larger language models don't care how you think: Why chain-of-thought prompting fails in subjective tasks.
\newblock In \emph{{{ICASSP}} 2025-2025 {{IEEE International Conference}} on {{Acoustics}}, {{Speech}} and {{Signal Processing}} ({{ICASSP}})}, pages 1--5. IEEE.

\bibitem[{Chochlakis et~al.(2024)Chochlakis, Potamianos, Lerman, and Narayanan}]{chochlakis2024strong}
Georgios Chochlakis, Alexandros Potamianos, Kristina Lerman, and Shrikanth Narayanan. 2024.
\newblock \href {https://arxiv.org/abs/2403.17125} {The strong pull of prior knowledge in large language models and its impact on emotion recognition}.
\newblock In \emph{2024 12th International Conference on Affective Computing and Intelligent Interaction (ACII)}. IEEE.

\bibitem[{Cowen and Keltner(2017)}]{cowenSelfreportCaptures272017}
Alan~S Cowen and Dacher Keltner. 2017.
\newblock Self-report captures 27 distinct categories of emotion bridged by continuous gradients.
\newblock 114(38):E7900--E7909.

\bibitem[{Davani et~al.(2022)Davani, Díaz, and Prabhakaran}]{davaniDealingDisagreementsLooking2022}
Aida~Mostafazadeh Davani, Mark Díaz, and Vinodkumar Prabhakaran. 2022.
\newblock Dealing with disagreements: {{Looking}} beyond the majority vote in subjective annotations.
\newblock 10:92--110.

\bibitem[{Dawid and Skene(1979)}]{dawidMaximumLikelihoodEstimation1979a}
Alexander~Philip Dawid and Allan~M Skene. 1979.
\newblock Maximum likelihood estimation of observer error-rates using the {{EM}} algorithm.
\newblock 28(1):20--28.

\bibitem[{Demszky et~al.(2020)Demszky, Movshovitz-Attias, Ko, Cowen, Nemade, and Ravi}]{demszky2020goemotions}
Dorottya Demszky, Dana Movshovitz-Attias, Jeongwoo Ko, Alan Cowen, Gaurav Nemade, and Sujith Ravi. 2020.
\newblock {{GoEmotions}}: A dataset of fine-grained emotions.
\newblock In \emph{Proceedings of the 58th Annual Meeting of the Association for Computational Linguistics}, pages 4040--4054.

\bibitem[{Devlin et~al.(2018)Devlin, Chang, Lee, and Toutanova}]{devlin2018bert}
Jacob Devlin, Ming-Wei Chang, Kenton Lee, and Kristina Toutanova. 2018.
\newblock \href {https://arxiv.org/abs/1810.04805} {Bert: {{Pre-training}} of deep bidirectional transformers for language understanding}.

\bibitem[{Dubey et~al.(2024)Dubey, Jauhri, Pandey, Kadian, Al-Dahle, Letman, Mathur, Schelten, Yang, Fan et~al.}]{dubey2024llama}
Abhimanyu Dubey, Abhinav Jauhri, Abhinav Pandey, Abhishek Kadian, Ahmad Al-Dahle, Aiesha Letman, Akhil Mathur, Alan Schelten, Amy Yang, Angela Fan, et~al. 2024.
\newblock The llama 3 herd of models.
\newblock \emph{arXiv preprint arXiv:2407.21783}.

\bibitem[{Dutta et~al.(2023)Dutta, Mittal, Chen, Ramachandran, Rajakumar, Kivlichan, Mak, Butryna, and Paritosh}]{dutta2023modeling}
Senjuti Dutta, Sid Mittal, Sherol Chen, Deepak Ramachandran, Ravi Rajakumar, Ian Kivlichan, Sunny Mak, Alena Butryna, and Praveen Paritosh. 2023.
\newblock \href {https://arxiv.org/abs/2311.00203} {Modeling subjectivity (by {{Mimicking Annotator Annotation}}) in toxic comment identification across diverse communities}.
\newblock \emph{Preprint}, arXiv:2311.00203.

\bibitem[{Ferrara(2023)}]{ferrara2023should}
Emilio Ferrara. 2023.
\newblock Should chatgpt be biased? challenges and risks of bias in large language models.
\newblock \emph{arXiv preprint arXiv:2304.03738}.

\bibitem[{Garten et~al.(2019)Garten, Kennedy, Hoover, Sagae, and Dehghani}]{gartenIncorporatingDemographicEmbeddings2019}
Justin Garten, Brendan Kennedy, Joe Hoover, Kenji Sagae, and Morteza Dehghani. 2019.
\newblock Incorporating demographic embeddings into language understanding.
\newblock 43(1):e12701.

\bibitem[{Golazizian et~al.(2024)Golazizian, Omrani, Ziabari, and Dehghani}]{golazizian2024cost}
Preni Golazizian, Ali Omrani, Alireza~S Ziabari, and Morteza Dehghani. 2024.
\newblock Cost-efficient subjective task annotation and modeling through few-shot annotator adaptation.
\newblock \emph{arXiv preprint arXiv:2402.14101}.

\bibitem[{Gonen and Goldberg(2019)}]{gonen2019lipstick}
Hila Gonen and Yoav Goldberg. 2019.
\newblock Lipstick on a pig: Debiasing methods cover up systematic gender biases in word embeddings but do not remove them.
\newblock \emph{arXiv preprint arXiv:1903.03862}.

\bibitem[{Gordon et~al.(2022)Gordon, Lam, Park, Patel, Hancock, Hashimoto, and Bernstein}]{gordonJuryLearningIntegrating2022}
Mitchell~L Gordon, Michelle~S Lam, Joon~Sung Park, Kayur Patel, Jeff Hancock, Tatsunori Hashimoto, and Michael~S Bernstein. 2022.
\newblock Jury learning: {{Integrating}} dissenting voices into machine learning models.
\newblock In \emph{Proceedings of the 2022 {{CHI Conference}} on {{Human Factors}} in {{Computing Systems}}}, pages 1--19.

\bibitem[{Gupta et~al.(2023)Gupta, Rosenbaum, and Elenberg}]{gupta2023gistscore}
Shivanshu Gupta, Clemens Rosenbaum, and Ethan~R Elenberg. 2023.
\newblock \href {https://arxiv.org/abs/2311.09606} {{{GistScore}}: {{Learning}} better representations for in-context example selection with gist bottlenecks}.

\bibitem[{Hartmann et~al.(2023)Hartmann, Schwenzow, and Witte}]{hartmann2023political}
Jochen Hartmann, Jasper Schwenzow, and Maximilian Witte. 2023.
\newblock The political ideology of conversational ai: Converging evidence on chatgpt’s pro-environmental, left-libertarian orientation.
\newblock \emph{Left-Libertarian Orientation (January 1, 2023)}.

\bibitem[{Hovy et~al.(2013)Hovy, Berg-Kirkpatrick, Vaswani, and Hovy}]{hovy2013learning}
Dirk Hovy, Taylor Berg-Kirkpatrick, Ashish Vaswani, and Eduard Hovy. 2013.
\newblock Learning whom to trust with {{MACE}}.
\newblock In \emph{Proceedings of the 2013 Conference of the North American Chapter of the Association for Computational Linguistics: {{Human}} Language Technologies}, pages 1120--1130.

\bibitem[{Kossen et~al.(2023)Kossen, Rainforth, and Gal}]{kossen2023context}
Jannik Kossen, Tom Rainforth, and Yarin Gal. 2023.
\newblock \href {https://arxiv.org/abs/2307.12375} {In-context learning in large language models learns label relationships but is not conventional learning}.

\bibitem[{Lanham et~al.(2023)Lanham, Chen, Radhakrishnan, Steiner, Denison, Hernandez, Li, Durmus, Hubinger, Kernion, Lukošiūtė, Nguyen, Cheng, Joseph, Schiefer, Rausch, Larson, McCandlish, Kundu, Kadavath, Yang, Henighan, Maxwell, Telleen-Lawton, Hume, Hatfield-Dodds, Kaplan, Brauner, Bowman, and Perez}]{lanham2023measuring}
Tamera Lanham, Anna Chen, Ansh Radhakrishnan, Benoit Steiner, Carson Denison, Danny Hernandez, Dustin Li, Esin Durmus, Evan Hubinger, Jackson Kernion, Kamilė Lukošiūtė, Karina Nguyen, Newton Cheng, Nicholas Joseph, Nicholas Schiefer, Oliver Rausch, Robin Larson, Sam McCandlish, Sandipan Kundu, Saurav Kadavath, Shannon Yang, Thomas Henighan, Timothy Maxwell, Timothy Telleen-Lawton, Tristan Hume, Zac Hatfield-Dodds, Jared Kaplan, Jan Brauner, Samuel~R. Bowman, and Ethan Perez. 2023.
\newblock \href {https://arxiv.org/abs/2307.13702} {Measuring faithfulness in chain-of-thought reasoning}.
\newblock \emph{Preprint}, arXiv:2307.13702.

\bibitem[{Liu et~al.(2024)Liu, Li, Wu, and Lee}]{liu2024visual}
Haotian Liu, Chunyuan Li, Qingyang Wu, and Yong~Jae Lee. 2024.
\newblock Visual instruction tuning.
\newblock \emph{Advances in neural information processing systems}, 36.

\bibitem[{Marchisio et~al.(2024)Marchisio, Dash, Chen, Aumiller, {\"U}st{\"u}n, Hooker, and Ruder}]{marchisio2024does}
Kelly Marchisio, Saurabh Dash, Hongyu Chen, Dennis Aumiller, Ahmet {\"U}st{\"u}n, Sara Hooker, and Sebastian Ruder. 2024.
\newblock How does quantization affect multilingual llms?
\newblock \emph{arXiv preprint arXiv:2407.03211}.

\bibitem[{Min et~al.(2022)Min, Lyu, Holtzman, Artetxe, Lewis, Hajishirzi, and Zettlemoyer}]{min2022rethinking}
Sewon Min, Xinxi Lyu, Ari Holtzman, Mikel Artetxe, Mike Lewis, Hannaneh Hajishirzi, and Luke Zettlemoyer. 2022.
\newblock Rethinking the role of demonstrations: {{What}} makes in-context learning work?
\newblock In \emph{Proceedings of the 2022 Conference on Empirical Methods in Natural Language Processing}, pages 11048--11064.

\bibitem[{Mohammad et~al.(2018)Mohammad, Bravo-Marquez, Salameh, and Kiritchenko}]{mohammad2018semeval}
Saif Mohammad, Felipe Bravo-Marquez, Mohammad Salameh, and Svetlana Kiritchenko. 2018.
\newblock Semeval-2018 task 1: {{Affect}} in tweets.
\newblock In \emph{Proceedings of the 12th International Workshop on Semantic Evaluation}, pages 1--17.

\bibitem[{Mokhberian et~al.(2022)Mokhberian, Hopp, Harandizadeh, Morstatter, and Lerman}]{mokhberian2022noise}
Negar Mokhberian, Frederic~R Hopp, Bahareh Harandizadeh, Fred Morstatter, and Kristina Lerman. 2022.
\newblock Noise audits improve moral foundation classification.
\newblock In \emph{2022 IEEE/ACM International Conference on Advances in Social Networks Analysis and Mining (ASONAM)}, pages 147--154. IEEE.

\bibitem[{Mokhberian et~al.(2023)Mokhberian, Marmarelis, Hopp, Basile, Morstatter, and Lerman}]{mokhberian2023capturing}
Negar Mokhberian, Myrl~G Marmarelis, Frederic~R Hopp, Valerio Basile, Fred Morstatter, and Kristina Lerman. 2023.
\newblock Capturing perspectives of crowdsourced annotators in subjective learning tasks.
\newblock \emph{arXiv preprint arXiv:2311.09743}.

\bibitem[{Ouyang et~al.(2022{\natexlab{a}})Ouyang, Wu, Jiang, Almeida, Wainwright, Mishkin, Zhang, Agarwal, Slama, Ray et~al.}]{ouyangTrainingLanguageModels2022}
Long Ouyang, Jeff Wu, Xu~Jiang, Diogo Almeida, Carroll~L Wainwright, Pamela Mishkin, Chong Zhang, Sandhini Agarwal, Katarina Slama, Alex Ray, et~al. 2022{\natexlab{a}}.
\newblock \href {https://arxiv.org/abs/2203.02155} {Training language models to follow instructions with human feedback}.

\bibitem[{Ouyang et~al.(2022{\natexlab{b}})Ouyang, Wu, Jiang, Almeida, Wainwright, Mishkin, Zhang, Agarwal, Slama, Ray et~al.}]{ouyang2022training}
Long Ouyang, Jeffrey Wu, Xu~Jiang, Diogo Almeida, Carroll Wainwright, Pamela Mishkin, Chong Zhang, Sandhini Agarwal, Katarina Slama, Alex Ray, et~al. 2022{\natexlab{b}}.
\newblock Training language models to follow instructions with human feedback.
\newblock \emph{Advances in neural information processing systems}, 35:27730--27744.

\bibitem[{Pan et~al.(2023)Pan, Gao, Chen, and Chen}]{pan2023context}
Jane Pan, Tianyu Gao, Howard Chen, and Danqi Chen. 2023.
\newblock \href {https://arxiv.org/abs/2305.09731} {What in-context learning" learns" in-context: {{Disentangling}} task recognition and task learning}.

\bibitem[{Prabhakaran et~al.(2021)Prabhakaran, Davani, and Diaz}]{prabhakaranReleasingAnnotatorlevelLabels2021}
Vinodkumar Prabhakaran, Aida~Mostafazadeh Davani, and Mark Diaz. 2021.
\newblock \href {https://arxiv.org/abs/2110.05699} {On releasing annotator-level labels and information in datasets}.

\bibitem[{Radford et~al.(2019)Radford, Wu, Child, Luan, Amodei, Sutskever et~al.}]{radford2019language}
Alec Radford, Jeffrey Wu, Rewon Child, David Luan, Dario Amodei, Ilya Sutskever, et~al. 2019.
\newblock Language models are unsupervised multitask learners.
\newblock 1(8):9.

\bibitem[{Resnick et~al.(2021)Resnick, Kong, Schoenebeck, and Weninger}]{resnick2021survey}
Paul Resnick, Yuqing Kong, Grant Schoenebeck, and Tim Weninger. 2021.
\newblock Survey equivalence: A procedure for measuring classifier accuracy against human labels.
\newblock \emph{arXiv preprint arXiv:2106.01254}.

\bibitem[{Rubin et~al.(2022)Rubin, Herzig, and Berant}]{rubin2022learning}
Ohad Rubin, Jonathan Herzig, and Jonathan Berant. 2022.
\newblock Learning to retrieve prompts for in-context learning.
\newblock In \emph{Proceedings of the 2022 Conference of the North American Chapter of the Association for Computational Linguistics: {{Human}} Language Technologies}, pages 2655--2671.

\bibitem[{Sap et~al.(2022)Sap, Swayamdipta, Vianna, Zhou, Choi, and Smith}]{sapAnnotatorsAttitudesHow2022}
Maarten Sap, Swabha Swayamdipta, Laura Vianna, Xuhui Zhou, Yejin Choi, and Noah~A. Smith. 2022.
\newblock \href {https://doi.org/10.18653/v1/2022.naacl-main.431} {Annotators with {{Attitudes}}: {{How Annotator Beliefs And Identities Bias Toxic Language Detection}}}.
\newblock In \emph{Proceedings of the 2022 {{Conference}} of the {{North American Chapter}} of the {{Association}} for {{Computational Linguistics}}: {{Human Language Technologies}}}, pages 5884--5906. Association for Computational Linguistics.

\bibitem[{Srivastava et~al.(2022)Srivastava, Rastogi, Rao, Shoeb, Abid, Fisch, Brown, Santoro, Gupta, Garriga-Alonso et~al.}]{srivastava2022beyond}
Aarohi Srivastava, Abhinav Rastogi, Abhishek Rao, Abu Awal~Md Shoeb, Abubakar Abid, Adam Fisch, Adam~R Brown, Adam Santoro, Aditya Gupta, Adrià Garriga-Alonso, et~al. 2022.
\newblock \href {https://arxiv.org/abs/2206.04615} {Beyond the imitation game: {{Quantifying}} and extrapolating the capabilities of language models}.

\bibitem[{Swayamdipta et~al.(2020)Swayamdipta, Schwartz, Lourie, Wang, Hajishirzi, Smith, and Choi}]{swayamdipta2020dataset}
Swabha Swayamdipta, Roy Schwartz, Nicholas Lourie, Yizhong Wang, Hannaneh Hajishirzi, Noah~A Smith, and Yejin Choi. 2020.
\newblock \href {https://arxiv.org/abs/2009.10795} {Dataset cartography: {{Mapping}} and diagnosing datasets with training dynamics}.

\bibitem[{Tak and Gratch(2024)}]{tak2024gpt}
Ala~N Tak and Jonathan Gratch. 2024.
\newblock Gpt-4 emulates average-human emotional cognition from a third-person perspective.
\newblock \emph{arXiv preprint arXiv:2408.13718}.

\bibitem[{Touvron et~al.(2023)Touvron, Martin, Stone, Albert, Almahairi, Babaei, Bashlykov, Batra, Bhargava, and Bhosale}]{touvronLlamaOpenFoundation2023}
Hugo Touvron, Louis Martin, Kevin Stone, Peter Albert, Amjad Almahairi, Yasmine Babaei, Nikolay Bashlykov, Soumya Batra, Prajjwal Bhargava, and Shruti Bhosale. 2023.
\newblock \href {https://arxiv.org/abs/2307.09288} {Llama 2: {{Open}} foundation and fine-tuned chat models}.

\bibitem[{Trager et~al.(2022)Trager, Ziabari, Davani, Golazizian, Karimi-Malekabadi, Omrani, Li, Kennedy, Reimer, Reyes et~al.}]{trager2022moral}
Jackson Trager, Alireza~S Ziabari, Aida~Mostafazadeh Davani, Preni Golazizian, Farzan Karimi-Malekabadi, Ali Omrani, Zhihe Li, Brendan Kennedy, Nils~Karl Reimer, Melissa Reyes, et~al. 2022.
\newblock \href {https://arxiv.org/abs/2208.05545} {The moral foundations reddit corpus}.

\bibitem[{Turpin et~al.(2024)Turpin, Michael, Perez, and Bowman}]{turpin2024language}
Miles Turpin, Julian Michael, Ethan Perez, and Samuel Bowman. 2024.
\newblock Language models don't always say what they think: unfaithful explanations in chain-of-thought prompting.
\newblock \emph{Advances in Neural Information Processing Systems}, 36.

\bibitem[{Vaswani et~al.(2017)Vaswani, Shazeer, Parmar, Uszkoreit, Jones, Gomez, Kaiser, and Polosukhin}]{vaswaniAttentionAllYou2017}
Ashish Vaswani, Noam Shazeer, Niki Parmar, Jakob Uszkoreit, Llion Jones, Aidan~N Gomez, \textbackslash~Lukasz Kaiser, and Illia Polosukhin. 2017.
\newblock Attention is all you need.
\newblock 30.

\bibitem[{Wang et~al.(2022)Wang, Wei, Schuurmans, Le, Chi, Narang, Chowdhery, and Zhou}]{wang2022self}
Xuezhi Wang, Jason Wei, Dale Schuurmans, Quoc Le, Ed~Chi, Sharan Narang, Aakanksha Chowdhery, and Denny Zhou. 2022.
\newblock Self-consistency improves chain of thought reasoning in language models.
\newblock \emph{arXiv preprint arXiv:2203.11171}.

\bibitem[{Wei et~al.(2022)Wei, Wang, Schuurmans, Bosma, Chi, Le, and Zhou}]{weiChainThoughtPrompting2022}
Jason Wei, Xuezhi Wang, Dale Schuurmans, Maarten Bosma, Ed~Chi, Quoc Le, and Denny Zhou. 2022.
\newblock \href {https://arxiv.org/abs/2201.11903} {Chain of thought prompting elicits reasoning in large language models}.

\bibitem[{Wei et~al.(2023)Wei, Wei, Tay, Tran, Webson, Lu, Chen, Liu, Huang, Zhou et~al.}]{wei2023larger}
Jerry Wei, Jason Wei, Yi~Tay, Dustin Tran, Albert Webson, Yifeng Lu, Xinyun Chen, Hanxiao Liu, Da~Huang, Denny Zhou, et~al. 2023.
\newblock \href {https://arxiv.org/abs/2303.03846} {Larger language models do in-context learning differently}.

\bibitem[{Wolf et~al.(2020)Wolf, Debut, Sanh, Chaumond, Delangue, Moi, Cistac, Rault, Louf, Funtowicz, Davison, Shleifer, von Platen, Ma, Jernite, Plu, Xu, Scao, Gugger, Drame, Lhoest, and Rush}]{wolf-etal-2020-transformers}
Thomas Wolf, Lysandre Debut, Victor Sanh, Julien Chaumond, Clement Delangue, Anthony Moi, Pierric Cistac, Tim Rault, Rémi Louf, Morgan Funtowicz, Joe Davison, Sam Shleifer, Patrick von Platen, Clara Ma, Yacine Jernite, Julien Plu, Canwen Xu, Teven~Le Scao, Sylvain Gugger, Mariama Drame, Quentin Lhoest, and Alexander~M. Rush. 2020.
\newblock \href {https://www.aclweb.org/anthology/2020.emnlp-demos.6} {Transformers: State-of-the-art natural language processing}.
\newblock In \emph{Proceedings of the 2020 Conference on Empirical Methods in Natural Language Processing: System Demonstrations}, pages 38--45, Online. Association for Computational Linguistics.

\bibitem[{Xie et~al.(2021)Xie, Raghunathan, Liang, and Ma}]{xie2021explanation}
Sang~Michael Xie, Aditi Raghunathan, Percy Liang, and Tengyu Ma. 2021.
\newblock An explanation of in-context learning as implicit bayesian inference.
\newblock In \emph{International Conference on Learning Representations}.

\bibitem[{Yang et~al.(2023)Yang, Kommineni, Alshehri, Mohanty, Modi, Gratch, and Narayanan}]{yang2023context}
Daniel Yang, Aditya Kommineni, Mohammad Alshehri, Nilamadhab Mohanty, Vedant Modi, Jonathan Gratch, and Shrikanth Narayanan. 2023.
\newblock Context unlocks emotions: Text-based emotion classification dataset auditing with large language models.
\newblock In \emph{2023 11th International Conference on Affective Computing and Intelligent Interaction (ACII)}, pages 1--8. IEEE.

\bibitem[{Yao et~al.(2024)Yao, Yu, Zhao, Shafran, Griffiths, Cao, and Narasimhan}]{yao2024tree}
Shunyu Yao, Dian Yu, Jeffrey Zhao, Izhak Shafran, Tom Griffiths, Yuan Cao, and Karthik Narasimhan. 2024.
\newblock Tree of thoughts: {{Deliberate}} problem solving with large language models.
\newblock 36.

\bibitem[{Yoo et~al.(2022)Yoo, Kim, Kim, Cho, Jo, Lee, Lee, and Kim}]{yoo2022ground}
Kang~Min Yoo, Junyeob Kim, Hyuhng~Joon Kim, Hyunsoo Cho, Hwiyeol Jo, Sang-Woo Lee, Sang-goo Lee, and Taeuk Kim. 2022.
\newblock Ground-truth labels matter: A deeper look into input-label demonstrations.
\newblock \emph{arXiv preprint arXiv:2205.12685}.

\end{thebibliography}

\appendix

\section{Annotators}

We reiterate that we group annotators together based on their overlap in the train and evaluation sets so as to standardize the prompt across them. We use 5 annotators in MFRC, one triplet and one pair, both with enough common examples in the evaluation set to present somewhat robust results. For both annotator groups, we use 100 evaluation examples. For GoEmotions, we use 6 annotators, as these were the only groups with enough evaluation examples, 100 and 71 respectively. We did not find any possible quadruplets or larger groups.

\section{Annotation Details} \label{sec:annotations}

\change{Two research assistants trained on emotions and moral foundations using the annotation manuals provided by the dataset authors and with common instructions independently generated the reasonings for the annotations in the datasets. We then check for the consistency of the reasoning chains between the two annotators qualitatively (manual checks) and quantitatively (checking the consistency of the predictions when using either annotator's reasonings). Note that we do not use model-generated CoT because we noticed from our experiments that LLM explanations are shallow and can easily miss more complex emotional and moral expressions.}

\section{Toy Example for Aggregation Artifact} \label{sec:app-inconsistency}

\change{In this section, we present an toy example to elucidate the notion of inconsistencies that can be created from aggregation, presented in the main text. Consider an example dataset where the US political affiliation is the major factor dictating the labels provided by annotators. Due to random assignment of examples to annotators, some examples may be assigned more annotators leaning Democrat, and similar for Republican. Therefore, the aggregated dataset will contain some annotations reflecting Republican views, and some Democrat views, creating inconsistencies in the training data, or worse, between training and test data. Critically, trying to model political affiliation, the major factor of this setting, becomes impossible from this aggregated data, hence the ``inconsistency'' of the annotations.}

\section{Prompt} \label{sec:prompt}

For completeness, we showcase example prompts we use in MFRC for ICL and CoT that illustrate the prompt format we utilize across all experiments. One example for each setting is shown in Table~\ref{tab:prompt-example}. We note that the prompts do not utilize the conversational format even though it is available across all our models because we found it to work worse in terms of performance (in terms of the performance of the aggregate in GoEmotions) and ability to decode (e.g., including explanations without being prompted to, diverging in terms of output format, or even predicting emojis rather than emotion words) compared to the used one.

\begin{table}[t]
    \centering
    \footnotesize
    \renewcommand{\arraystretch}{1}
    \begin{tabular}[t]{@{}p{23em} @{}}

    \textbf{Prompt examples} \\
    
    \midrule
    \textbf{ICL} \\ \\
    \narrowbotc{Classify the following inputs into none, one, or multiple the following moral foundations per input: authority, care, equality, loyalty, proportionality and purity.\\\\Input: Did anyone watch Nigel Farage rebuke the EU? It was quite interesting!\\Moral Foundation(s): none\\\\Input: Because Le Pen is alt right and is very dangerous for the peace of europe and the world.\\Moral Foundation(s):} \\ \\

    \textbf{CoT}  \\\\
    
    \narrowbotc{Classify the following inputs into none, one, or multiple the following moral foundations per input: authority, care, equality, loyalty, proportionality and purity.\\\\Input: Did anyone watch Nigel Farage rebuke the EU? It was quite interesting!\\\\Reasoning: The author expresses interest in Farage's actions but not any moral sentiment towards it.\\\\Moral Foundation(s): none\\\\Input: Because Le Pen is alt right and is very dangerous for the peace of europe and the world.\\\\Reasoning:}
    \end{tabular}
    \caption{Examples showcasing the prompt format in In-context Learning and Chain-of-Thought prompting.}
    \label{tab:prompt-example}
\end{table}

\section{Quantization}

\change{We perform experiments with LLaMA-3 8B with less quantization to gauge at the effects that it has on our results. We present our findings in Table~\ref{tab:quant}. As expected, performance slightly improves with less quantization for our metrics of interest, yet the comparative phenomena we study still seem to hold from this small-scale analysis.}

\begin{table}
    \centering
    \begin{tabular}{lccc}
        \textbf{Q.} & \textbf{Micro F1} & \textbf{Macro F1} & \textbf{JS} \\
        \midrule
        \multicolumn{4}{c}{Aggregate} \\
        \midrule
        4 bit & \rescell{0.078}{0.026} & \rescell{0.094}{0.034} & \rescell{0.108}{0.013} \\
        8 bit & \rescell{0.124}{0.036} & \rescell{0.137}{0.017} & \rescell{0.241}{0.071} \\
        16 bit & \rescell{0.131}{0.047} & \rescell{0.150}{0.014} & \rescell{0.219}{0.061} \\
        \midrule
        \multicolumn{4}{c}{Annotator \#1} \\
        \midrule
        4 bit & \rescell{0.090}{0.030} & \rescell{0.111}{0.033} & \rescell{0.114}{0.046} \\
        8 bit & \rescell{0.123}{0.061} & \rescell{0.155}{0.049} & \rescell{0.240}{0.108} \\
        \midrule
        \multicolumn{4}{c}{Annotator \#2} \\
        \midrule
        4 bit & \rescell{0.243}{0.038} & \rescell{0.170}{0.024} & \rescell{0.245}{0.038} \\
        8 bit & \rescell{0.208}{0.059} & \rescell{0.228}{0.022} & \rescell{0.312}{0.042} \\
        \midrule
        \multicolumn{4}{c}{Annotator \#3} \\
        \midrule
        4 bit & \rescell{0.122}{0.043} & \rescell{0.122}{0.036} & \rescell{0.154}{0.090} \\
        8 bit & \rescell{0.126}{0.018} & \rescell{0.135}{0.020} & \rescell{0.247}{0.072} \\
        
    \end{tabular}
    \caption{\change{Performance of LLaMA-3 8B with various levels of quantization (Q.) on MFRC.}}
    \label{tab:quant}
\end{table}

\section{Full Results}

\begin{figure*}[t]
    \centering
  \includegraphics[width=\linewidth]{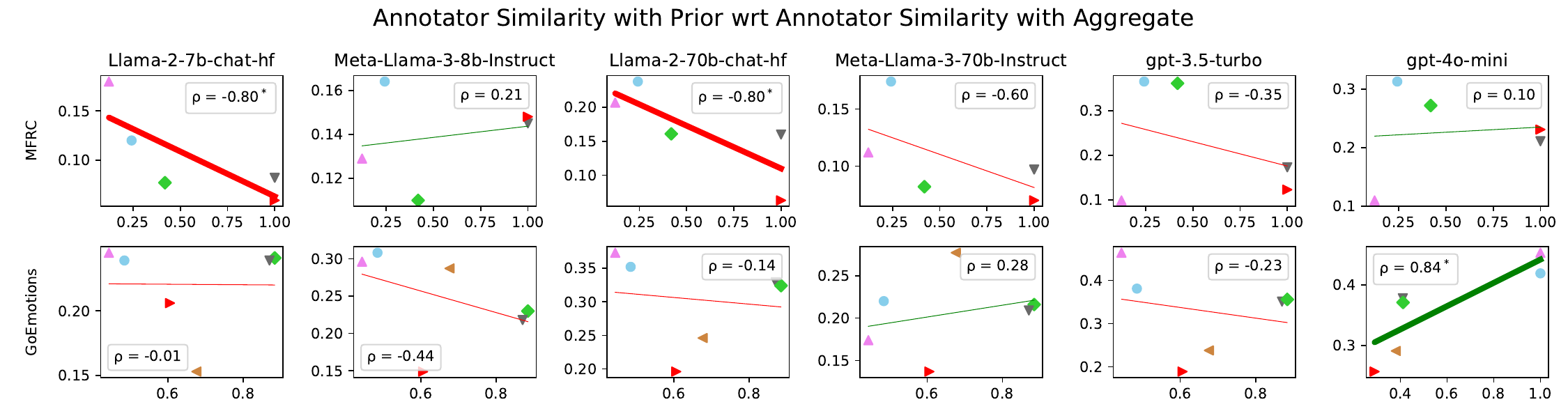}
  \caption{\change{Scatter plot of annotator similarity with aggregate and with prior, with correlation and line fit of data shown. $^*$ and bold lines: $p < 0.05$.}}
  \label{fig:prior-aggr-full}
\end{figure*}

\begin{figure*}[t]
    \centering
  \includegraphics[width=\linewidth]{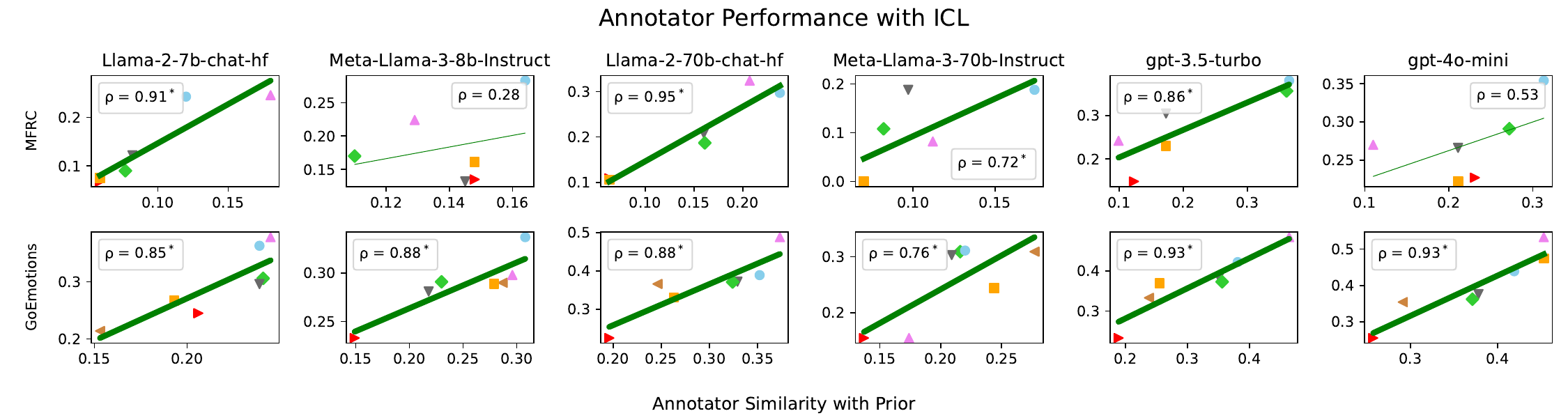}
  \includegraphics[width=\linewidth]{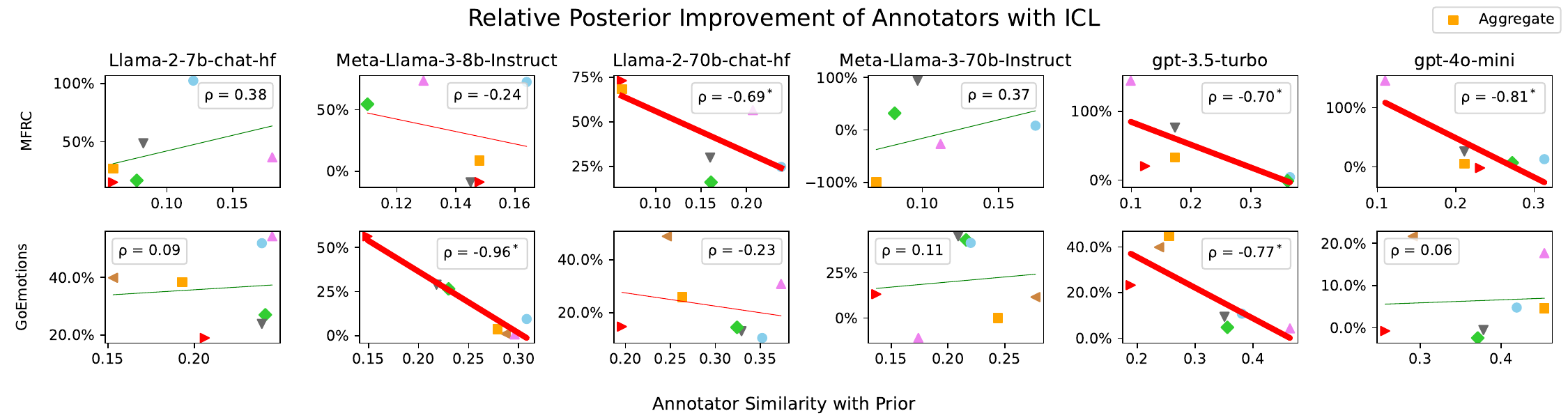}
  \caption {\change{\textbf{In-Context Learning} performance for annotators and aggregate and their relative improvement compared to the model's prior as a function of the similarity of each with the model's prior. Correlation and line fit of data also shown. $^*$ and bold lines: $p < 0.05$.}}
  \label{fig:prior-full}
\end{figure*}

\begin{figure*}[!]
  \centering
  \includegraphics[width=\linewidth]{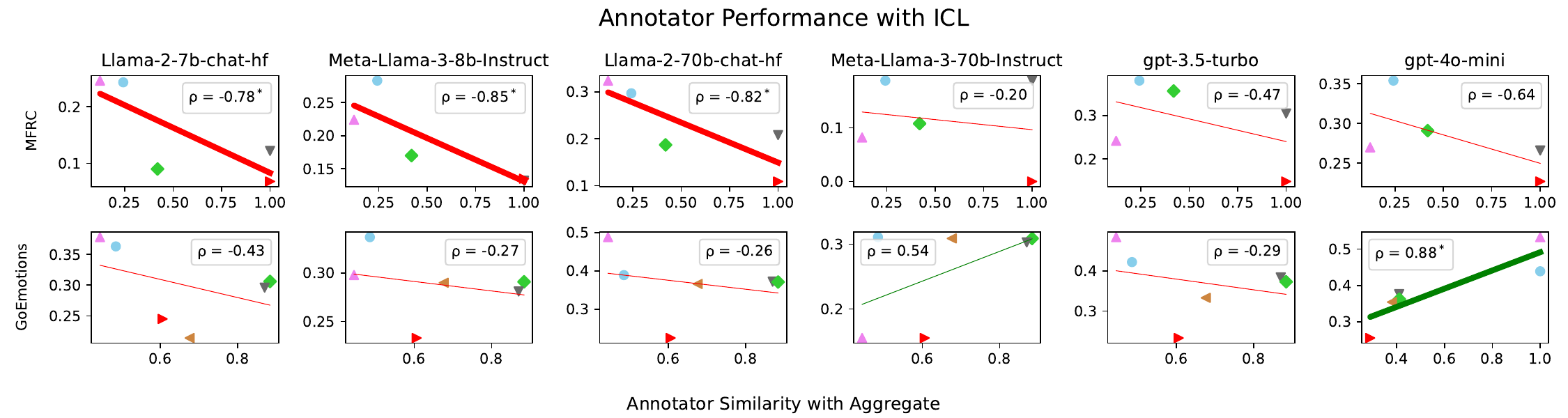}
  \includegraphics[width=\linewidth]{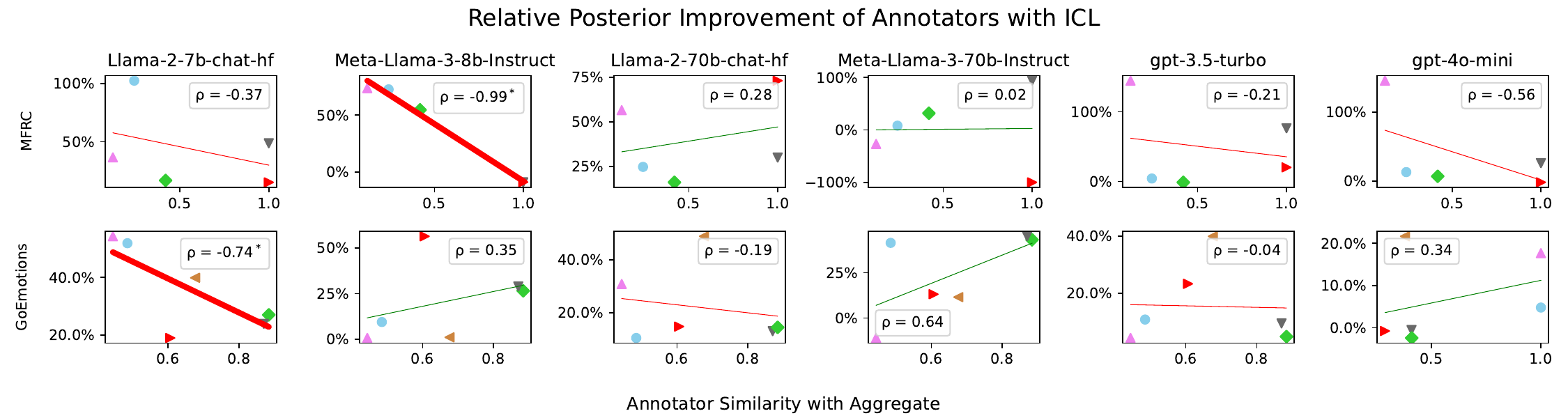}
  \caption {\change{\textbf{In-Context Learning} performance for annotators and their relative improvement compared to the model's prior as a function of the similarity of each annotator with the aggregate. Correlation and line fit of data also shown. $^*$ and bold lines: $p < 0.05$.}}
  \label{fig:aggr-full}
\end{figure*}

\begin{figure*}[t]
  \centering
  \includegraphics[width=\linewidth]{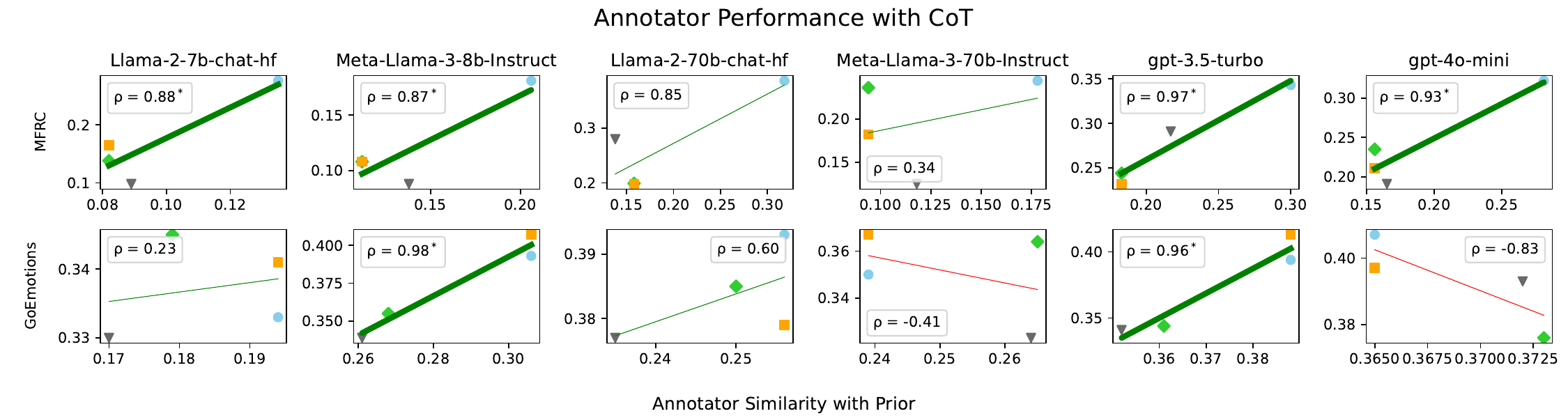}
  \includegraphics[width=\linewidth]{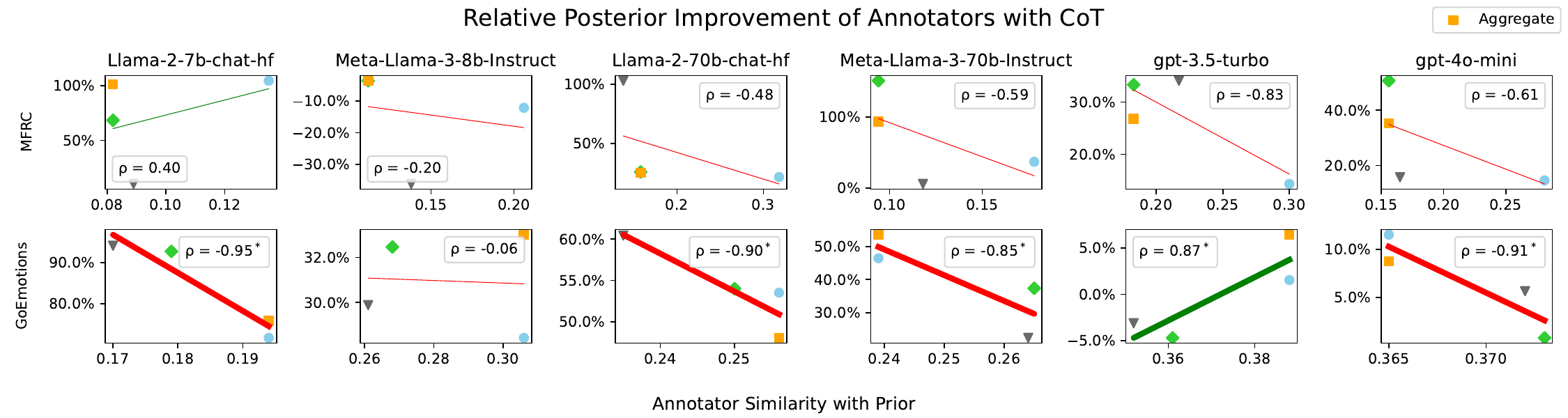}
  \caption {\change{\textbf{Chain-of-Thought} performance for annotators and aggregate and their relative improvement compared to the model's prior as a function of the similarity of each with the model's prior. Correlation and line fit of data also shown. $^*$ and bold lines: $p < 0.05$.}}
  \label{fig:prior-cot-full}
\end{figure*}

\begin{figure*}[t]
  \includegraphics[width=\linewidth]{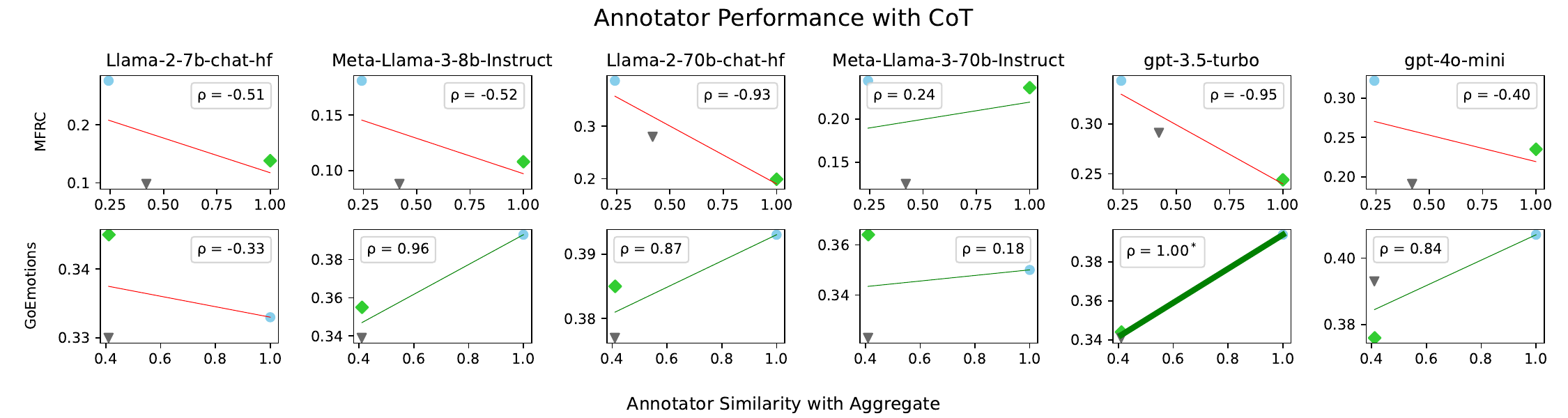}
  \includegraphics[width=\linewidth]{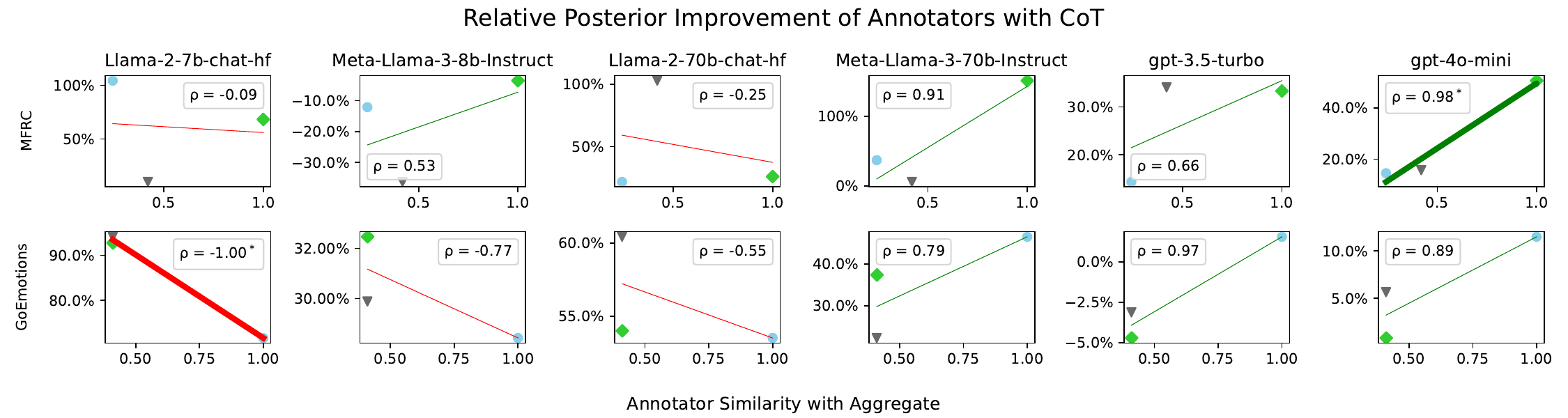}
  \caption{\change{\textbf{Chain-of-Thought} performance for annotators and their relative improvement compared to the model's prior as a function of the similarity of each annotator with the aggregate. Correlation and line fit of data also shown. $^*$ and bold lines: $p < 0.05$.}}
  \label{fig:aggr-cot-full}
\end{figure*}

\change{In Figures~\ref{fig:prior-aggr-full}, \ref{fig:prior-full}, \ref{fig:aggr-full}, \ref{fig:prior-cot-full}, and \ref{fig:aggr-cot-full}, we see the equivalent results to Figures~\ref{fig:prior-aggr}, \ref{fig:prior}, \ref{fig:aggr}, \ref{fig:prior-cot}, and \ref{fig:aggr-cot} respectively, but each annotator is assigned their own marker. Therefore, these more detailed figures allow us to track the performance of different annotators across models and settings.}

\end{document}